\documentclass[10pt,twocolumn,letterpaper]{article}
\usepackage{cvpr}
\usepackage{amsmath,amssymb,bm}
\usepackage{booktabs}
\usepackage{graphicx}
\usepackage{xcolor}
\usepackage{xspace}
\usepackage{multirow}
\usepackage{array}
\usepackage{tabularx}
\usepackage{enumitem}
\usepackage{microtype}
\usepackage{placeins}
\usepackage{url}
\usepackage{algorithm}
\usepackage{algpseudocode}

\definecolor{cvprblue}{rgb}{0.21,0.49,0.74}
\usepackage[pagebackref,breaklinks,colorlinks,allcolors=cvprblue]{hyperref}

\def\paperID{*****}
\def\confName{CVPR}
\def\confYear{2027}

\graphicspath{{figures/}{Figures/}}

\newcommand{\ours}{\textsc{OCL}\xspace}
\newcommand{\R}{\mathcal{R}}
\newcommand{\C}{\mathcal{C}}
\newcommand{\Oset}{\mathcal{O}}
\newcommand{\Aset}{\mathcal{A}}
\newcommand{\Mmem}{\mathcal{M}}
\newcommand{\Norm}{\operatorname{Norm}}
\newcommand{\sg}{\operatorname{sg}}

\newcommand{\best}[1]{\textcolor{red}{\textbf{#1}}}
\newcommand{\second}[1]{\textcolor{blue}{\textbf{#1}}}

\title{Satellite-Free Training for Drone-View Geo-Localization}

\author{
Tao Liu$^{1}$ \quad
Yingzhi Zhang$^{2}$ \quad
Xingrun Wang$^{1}$ \quad
Kan Ren$^{1}$ \quad
Xiaoqi Zhao$^{3}$ \\[0.6em]
$^{1}$Nanjing University of Science and Technology, China \\
$^{2}$University of Tsukuba, Japan \\
$^{3}$Nanyang Technological University, Singapore \\[0.5em]
{\tt\small
\{liutao23, 924116850233, k.ren\}@njust.edu.cn} \\
{\tt\small
zhang\_y@css.risk.tsukuba.ac.jp \quad
xiaoqi.zhao@ntu.edu.sg}
}
\begin{document}
\maketitle


\begin{abstract}
Drone-view geo-localization (DVGL) typically assumes access to satellite imagery during task-specific development. We study a late-arriving-gallery setting, where model development relies only on grouped source UAV observations and the target satellite gallery remains unavailable until the model is frozen. We propose \ours, Orbit Completion Learning, which uses source-side multi-view geometry as privileged training information while preserving image-only deployment. OCL reconstructs each source scene using structure-from-motion and 3D Gaussian Splatting, and renders canonical complementary views to complete its observation orbit. An exponential-moving-average teacher aggregates the completed orbit into a scene-level target, while a variable-cardinality student learns to recover this target from partial observations through distillation and contrastive learning. At inference, the frozen student independently encodes UAV queries and satellite gallery images using the same single-image path, without reconstruction or target-gallery adaptation. Experiments on University-1652, SUES-200, and IRVL328, together with nine weather-style appearance shifts, evaluate generalization across altitude, modality, and appearance variations. OCL achieves 57.85\% R@1 and 68.44\% mAP on University-1652, and obtains the best R@1 among the evaluated target-satellite-free methods in eight of nine weather conditions. Controlled ablations further show that completed teacher context provides gains beyond student-side synthetic-view augmentation. The code and data are publicly available at \url{https://github.com/liutao23/OCL-DVGL}.
\end{abstract}

\section{Introduction}
\label{sec:introduction}

\begin{figure*}[t]
    \centering
    \includegraphics[width=0.96\linewidth]{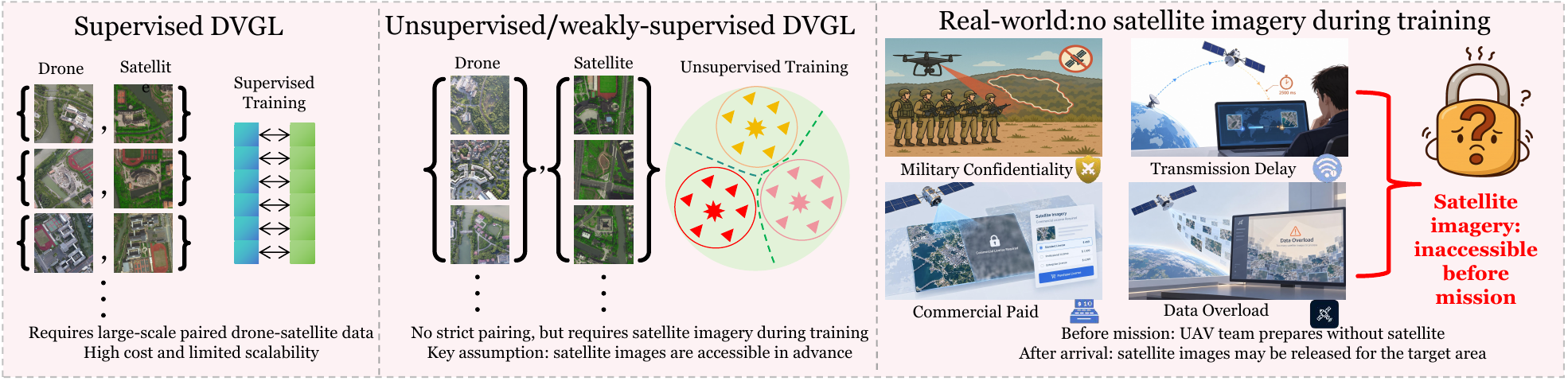}
    \caption{Existing DVGL methods require satellite imagery during training, whereas in real-world scenarios it may only become available after deployment. We therefore study DVGL using only source UAV observations during model development.}
    \label{fig:overview}
\end{figure*}

Drone-view geo-localization (DVGL) localizes a UAV by retrieving the corresponding geotagged satellite tile from a reference gallery. Existing supervised methods~\cite{wang2021each,dai2021transformer,deuser2023sample4geo,wang2024mfrgn,wen2025WeatherPrompt} and reduced-supervision approaches~\cite{li2024learning,li2024unleashing,chen2025limited,chen2026without} typically assume that satellite imagery is available during task-specific development, whether for paired supervision, pseudo-label generation, clustering, cross-view alignment, or representation fitting. In practice, however, satellite imagery for a target area may be unavailable during development because of access restrictions, licensing cost, transmission or release delay, or large-scale data-delivery constraints, even when source UAV observations are already available, as shown in Fig.~\ref{fig:overview}.

We adopt a single protocol boundary based on satellite access: whether target-benchmark satellite imagery or satellite-derived statistics are available before final retrieval. We study the \emph{target-satellite-free} setting, where task-specific development relies only on source UAV observations and the target satellite gallery is first accessed for descriptor extraction and retrieval. OCL is selected and frozen entirely on the source side; at test time, it independently encodes each UAV query and satellite image with the same single-image encoder, without query- or gallery-side fitting. This setting focuses on learning transferable place representations before the target satellite gallery becomes available.

A natural way to exploit multi-view UAV observations is to reconstruct the scene and render an overhead or bird's-eye-view proxy for cross-view matching~\cite{li2025unsupervised,Ju_2025_ICCV}. However, representing an entire UAV trajectory with a single synthesized view can discard complementary cues observed from different viewpoints. OCL uses reconstruction differently. Rather than generating a proxy for retrieval, we use source-side geometry only during training to render additional viewpoints and enrich the observations available for representation learning. At deployment, these geometric components are discarded, and the model localizes directly from the original UAV image.

Our \emph{Orbit Completion Learning} (OCL) framework reconstructs each source scene and renders canonical complementary views to complete its observation orbit. An EMA teacher summarizes the completed orbit into a scene-level target, while a variable-cardinality student learns to recover this target from partial observations. The geometric reconstruction and teacher are used only during training; at inference, only the student encoder is retained for image-based retrieval.

Our contributions are summarized as follows:
\begin{enumerate}[leftmargin=*,topsep=2pt,itemsep=1pt]
    \item To our knowledge, we are the first to formulate drone-view geo-localization without access to target satellite imagery during task-specific development. We study a \emph{late-arriving-gallery} setting in which the model is developed solely from source UAV observations and the target satellite gallery becomes available only at final retrieval.

    \item We propose \emph{Orbit Completion Learning} (OCL), which exploits source-side multi-view geometry as privileged training information. OCL completes the source observation orbit with geometrically consistent rendered views and distills this richer context into a student that operates directly on partial, and often single-image, observations.

    \item We evaluate OCL across multiple DVGL benchmarks covering altitude, modality, and appearance variations. Controlled ablations separate the effects of student-side synthetic augmentation and teacher-side orbit completion, while weather-shift experiments further assess robustness under challenging appearance changes.
\end{enumerate}

\section{Related Work}
\label{sec:related}

\subsection{Drone--Satellite Geo-Localization}
University-1652 established drone-to-satellite retrieval as a standard DVGL benchmark~\cite{zheng2020university}, followed by datasets studying altitude variation~\cite{zhu2023sues}, multi-scale and multi-temporal imagery~\cite{dai2023vision}, and infrared-to-visible localization~\cite{liu2025object}. Representative supervised methods include Instance Loss~\cite{zheng2020dual}, LPN~\cite{wang2021each}, FSRA~\cite{dai2021transformer}, Sample4Geo~\cite{deuser2023sample4geo}, Safe-Net~\cite{lin2024self}, MFRGN~\cite{wang2024mfrgn}, WeatherPrompt~\cite{wen2025WeatherPrompt}, and DiffusionUavLoc~\cite{liu2026diffusionuavloc}. Despite different architectures and learning strategies, these methods rely on satellite imagery during task-specific development. In contrast, our setting excludes target satellite imagery from training, validation, hyperparameter selection, and model selection.

\subsection{Reduced-Supervision Geo-Localization}
Recent methods reduce dependence on paired drone--satellite annotations through pseudo-labeling, clustering, weak supervision, or self-supervised alignment~\cite{li2024learning,li2024unleashing,chen2025limited,wang2025coarse,chen2026without}. However, they still access satellite imagery during model development. They therefore relax \emph{pairwise supervision}, rather than remove access to the target satellite domain. Our target-satellite-free setting instead develops the model solely from source UAV observations before the satellite gallery becomes available.

\subsection{Place Representation and Geometry}
Visual place recognition methods learn compact global descriptors through feature aggregation, including NetVLAD~\cite{arandjelovic2016netvlad}, SALAD~\cite{li2025place}, BoQ~\cite{ali2024boq}, and AnyLoc~\cite{keetha2023anyloc}. Set Transformer~\cite{lee2019settransformer} and Perceiver~\cite{jaegle2021perceiver} further provide permutation-invariant aggregation mechanisms. Separately, 3D Gaussian Splatting enables efficient novel-view rendering~\cite{kerbl3Dgaussians}, and recent geo-localization methods use reconstruction to synthesize normalized overhead views for matching~\cite{li2025unsupervised,Ju_2025_ICCV}. OCL uses geometry differently: reconstruction is used only during source-side training to generate complementary viewpoints, which provide privileged supervision for learning an image-based place representation. At inference, all geometric components are discarded and the frozen student directly encodes UAV and satellite images.

\section{Method}
\label{sec:method}

\begin{figure*}[t]
    \centering
    \includegraphics[width=\textwidth]{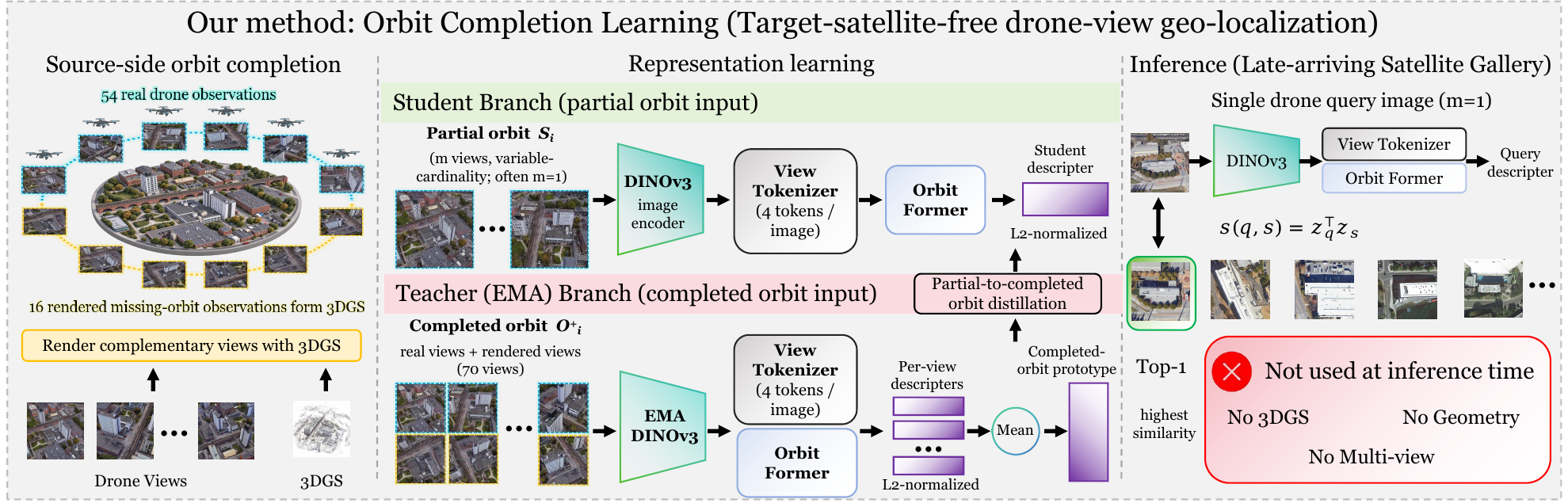}
    \caption{Overview of Orbit Completion Learning (OCL). OCL completes the source observation orbit with 3DGS-rendered views, distills completed-orbit context into a partial-orbit student, and performs single-image retrieval without geometry at inference.}
    \label{fig:method_overview}
\end{figure*}

Fig.~\ref{fig:method_overview} presents an overview of OCL. It consists of source-side orbit completion, partial-to-completed representation learning, and image-only retrieval at inference.

\subsection{Problem Setting}
Let scene $i$ contain a set of grouped source UAV observations
\begin{equation}
    \R_i=\{I^r_{i,n}\}_{n=1}^{n_i},
\end{equation}
where the grouping is given by the source location identities provided by the benchmark.
During task-specific development, OCL uses only these source UAV observations; target satellite imagery is not used for supervision or model selection.

From $\R_i$, we reconstruct the source scene and render a set of complementary views
\begin{equation}
    \C_i=\{I^c_{i,k}\}_{k=1}^{K_c}.
\end{equation}
Together, the observed and rendered views form the completed observation orbit
\begin{equation}
    \Oset_i=\R_i\cup\C_i.
\end{equation}
During training, the student receives a sampled partial observation set
$\Aset_i\subseteq\Oset_i$, whose cardinality varies across iterations.
The student encoder $f_\theta$ maps $\Aset_i$ to an
L2-normalized place descriptor $\bm z_i\in\mathbb{R}^{768}$.
An EMA teacher $f_{\bar\theta}$ uses the completed orbit to provide a scene-level learning target.

\subsection{Geometry-Consistent Orbit Completion}
\label{sec:geometry}

The purpose of orbit completion is to expose the learner to viewpoints that are not present in the observed UAV trajectory while preserving the geometry of the source scene.
For each scene, COLMAP estimates camera intrinsics, extrinsics, and sparse structure from the overlapping UAV observations~\cite{schoenberger2016sfm}.
The recovered cameras and sparse points initialize a 3D Gaussian Splatting (3DGS) model~\cite{kerbl3Dgaussians}.
Because the reconstruction coordinate system is defined only up to a global similarity transform, neither absolute georeferencing nor metric scale is required.

We represent scene $i$ by
\begin{equation}
    \mathcal G_i=
    \{(\bm\mu_g,\bm\Sigma_g,\alpha_g,\bm a_g)\}_{g=1}^{G_i},
\end{equation}
and optimize it using only the source UAV images with the standard photometric objective
\begin{equation}
    \mathcal L_{\mathrm{GS}}
    =
    0.8\|\hat I-I\|_1
    +0.2\bigl(1-\mathrm{SSIM}(\hat I,I)\bigr).
\end{equation}
This reconstruction stage is optimized independently from representation learning and is used only to generate additional training observations.

\textbf{Canonical orbit construction.}
To render comparable complementary views across scenes, we first construct a canonical camera orbit in the local reconstruction frame.
Let $\bm c_{i,n}$ denote the recovered center of camera $n$ and
$\bm d_{i,n}$ its unit viewing direction.
We estimate a scene target $\bm t_i$ as the robust intersection of the recovered viewing rays:
\begin{equation}
    \bm t_i=
    \arg\min_{\bm t}
    \sum_{n\in\mathcal J_i}
    w_{i,n}
    \left\|
    (\bm I-\bm d_{i,n}\bm d_{i,n}^{\top})
    (\bm t-\bm c_{i,n})
    \right\|_2^2.
    \label{eq:target}
\end{equation}
The weights $w_{i,n}$ are updated using IRLS to reduce the influence of poorly estimated viewing rays.

We then estimate the dominant orbit plane from the recovered camera centers.
The eigenvector associated with the smallest eigenvalue of their centered covariance defines the orbit normal $\bm u_i$.
Projecting a deterministic radial direction onto this plane gives the first in-plane axis $\bm e_{i,1}$, and we define
$\bm e_{i,2}=\bm u_i\times\bm e_{i,1}$.
The characteristic orbit radius is
\begin{equation}
    \rho_i=
    \operatorname{median}_{n\in\mathcal J_i}
    \|\bm c_{i,n}-\bm t_i\|_2.
\end{equation}

Given azimuth $\theta$ and tilt $\phi$, we define
\begin{align}
    \bm h_i(\theta)
    &=
    \sin\theta\,\bm e_{i,1}
    +\cos\theta\,\bm e_{i,2},\\
    \bm f_i(\theta,\phi)
    &=
    \sin\phi\,\bm h_i(\theta)
    -\cos\phi\,\bm u_i,\\
    \bm c_i^c(\theta,\phi)
    &=
    \bm t_i-\rho_i\bm f_i(\theta,\phi).
\end{align}
We use four azimuths
$\{0^\circ,90^\circ,180^\circ,270^\circ\}$
and four tilts
$\{35^\circ,25^\circ,15^\circ,5^\circ\}$,
where tilt is measured from the downward-looking direction.
This produces $K_c=16$ canonical complementary views per scene.

Each rendered camera is oriented toward $\bm t_i$, while its image resolution and horizontal and vertical fields of view are inherited from the median statistics of the reconstructed source cameras.
Because all quantities are defined within the reconstructed scene, the canonical orbit is unaffected by the arbitrary global translation, rotation, and scale of the SfM coordinate system.
Additional details on IRLS, sign disambiguation, and camera conventions are provided in the supplementary material.

\subsection{Variable-Cardinality Place Encoder}

OCL must learn from multiple observations during training while retaining the single-image operating point required by standard drone-to-satellite retrieval.
We therefore design the encoder to accept an unordered set of arbitrary cardinality.

Given one $256\times256$ image, a DINOv3 ViT-B/16 backbone~\cite{simeoni2025dinov3} produces patch tokens
$\bm X\in\mathbb R^{256\times768}$.
A \emph{ViewTokenizer} summarizes each image into four learned tokens.
Starting from learned queries
$\bm V^{(0)}\in\mathbb R^{4\times768}$,
two cross-attention blocks update them as
\begin{align}
    \widetilde{\bm V}^{(\ell)}
    &=
    \bm V^{(\ell-1)}
    +
    \mathrm{MHA}\!\left(
    \mathrm{LN}(\bm V^{(\ell-1)}),
    \mathrm{LN}(\bm X),
    \mathrm{LN}(\bm X)
    \right),\\
    \bm V^{(\ell)}
    &=
    \widetilde{\bm V}^{(\ell)}
    +
    \mathrm{FFN}\!\left(
    \mathrm{LN}(\widetilde{\bm V}^{(\ell)})
    \right).
\end{align}

For an input set containing $m$ observations, their summary tokens are concatenated into
$\bm U\in\mathbb R^{4m\times768}$.
The subsequent \emph{OrbitFormer} maintains one PLACE latent and eight auxiliary latents,
\begin{equation}
    \bm Z^{(0)}
    =
    [\bm z_{\mathrm{place}};\bm z_1;\ldots;\bm z_8].
\end{equation}
The latents first attend to $\bm U$ through latent-to-view cross-attention and are then refined by three self-attention/FFN blocks.
The L2-normalized PLACE latent is used as the final place descriptor.

We deliberately omit observation-index and camera-pose embeddings.
Consequently, permuting the input observations only permutes the corresponding key/value tokens and does not change the resulting descriptor.
The exact normalization scheme and a formal permutation-invariance derivation are given in the supplementary material.

\subsection{Completed-Orbit Teacher Target}

The central idea of OCL is to use the completed observation orbit as privileged context during training.
Rather than requiring the student to reproduce individual rendered views, we summarize the entire orbit into a scene-level representation and use it as the learning target.

The teacher is initialized from the student and updated by exponential moving average after each successful optimization step:
\begin{equation}
    \bar\theta_{t+1}
    =
    \alpha_t\bar\theta_t
    +(1-\alpha_t)\theta_{t+1},
\end{equation}
where $\alpha_t$ follows a cosine schedule from $0.996$ to $0.9999$.

Each observation in the completed orbit is independently encoded through the teacher's single-image path:
\begin{equation}
    \bar{\bm z}_{i,v}
    =
    f_{\bar\theta}(\{I_{i,v}\}),
    \qquad
    \|\bar{\bm z}_{i,v}\|_2=1.
\end{equation}
We then define the completed-orbit prototype as the normalized mean of these per-view descriptors:
\begin{equation}
    \bm c_i=
    \sg\!\left[
    \Norm\!\left(
    \frac{1}{|\Oset_i|}
    \sum_{I_{i,v}\in\Oset_i}
    \bar{\bm z}_{i,v}
    \right)
    \right].
    \label{eq:prototype}
\end{equation}
The resulting target is permutation invariant and introduces no additional learnable aggregation parameters.

Computing Eq.~\eqref{eq:prototype} from all images of every scene at every iteration would require repeated full-orbit backbone evaluation.
To avoid this cost, we maintain a cache of EMA-teacher ViewTokenizer outputs for individual observations.
At the beginning of each epoch, a rotating 20\% subset of the cached observations is re-encoded with the current teacher.
All scene prototypes are then rebuilt using the current EMA OrbitFormer and kept fixed, with gradients stopped, for the remainder of the epoch.
Cached entries corresponding to sampled observations are additionally updated by momentum during training.
This design makes completed-orbit supervision compatible with all-scene negatives without repeatedly forwarding every orbit through the image backbone.

\subsection{Partial-to-Completed Learning}

Given a sampled partial observation set $\Aset_i$, the student produces
\begin{equation}
    \bm z_i=f_\theta(\Aset_i).
\end{equation}
Training encourages $\bm z_i$ to recover the representation of the completed orbit in two complementary ways.

First, cosine distillation directly aligns the partial-orbit student descriptor with its completed-orbit teacher target:
\begin{equation}
    \mathcal L_{\mathrm{distill}}
    =
    \frac{1}{B}
    \sum_{i=1}^{B}
    \left(
    1-\bm z_i^\top\bm c_i
    \right).
\end{equation}
Second, an all-scene InfoNCE objective preserves place discrimination by contrasting each student descriptor against the prototypes of all training scenes:
\begin{equation}
    \mathcal L_{\mathrm{NCE}}
    =
    -\frac{1}{B}
    \sum_{i=1}^{B}
    \log
    \frac{
    \exp(\bm z_i^\top\bm c_i/\tau)
    }{
    \sum_{j\in\mathcal I_{\mathrm{tr}}}
    \exp(\bm z_i^\top\bm c_j/\tau)
    },
    \qquad
    \tau=0.07.
\end{equation}
The final objective is
\begin{equation}
    \mathcal L
    =
    \mathcal L_{\mathrm{distill}}
    +
    \mathcal L_{\mathrm{NCE}}.
\end{equation}\\
\textbf{Partial-orbit curriculum.}
During training, we progressively reduce the number of observations presented to the student.
Early iterations emphasize larger observation sets, allowing the model to learn from richer multi-view context, whereas later iterations increasingly emphasize the single-image case used at inference.

\begin{table}[t]
\centering
\scriptsize
\setlength{\tabcolsep}{3.8pt}
\begin{tabular}{lccccc}
\toprule
Progress & $m{=}16$ & $m{=}8$ & $m{=}4$ & $m{=}2$ & $m{=}1$\\
\midrule
0--25\%   & .40 & .35 & .15 & .05 & .05\\
25--60\%  & --  & .25 & .30 & .25 & .20\\
60--100\% & --  & --  & .15 & .30 & .55\\
\bottomrule
\end{tabular}
\caption{Observation-count curriculum. In the final 40\% of training, at least half of the scene samples in each logical batch are additionally forced to use a single observation.}
\label{tab:curriculum_main}
\end{table}

For each training sample, the student draws from the observed-only pool with probability $0.7$ and from the combined observed-and-rendered pool with probability $0.3$.
When $m>1$, we ensure that at least one real UAV observation is included.
This sampling strategy exposes the student to complementary rendered views while keeping the training distribution anchored to real UAV imagery.\\
\textbf{Inference.}
After source-side model selection, the student encoder is frozen and all training-time geometric components are discarded.
Each UAV query and each satellite gallery image is independently encoded using the same single-image ($m=1$) path.
Retrieval is performed directly by cosine similarity between their L2-normalized descriptors.
No reconstruction, rendered views, camera poses, teacher targets, or prototype memory are required at inference.

\section{Experiments}
\label{sec:experiments}

\subsection{Protocol and Implementation}

\textbf{Datasets.}
We evaluate on University-1652, SUES-200, and IRVL328. Following~\cite{liu2025object}, models are trained and selected on the University-1652 training split and evaluated on its test split as well as SUES-200 and IRVL328 without further adaptation. University-1652~\cite{zheng2020university} contains multi-view UAV observations from 1,652 locations with disjoint training and test splits. SUES-200~\cite{zhu2023sues} evaluates drone-to-satellite retrieval at four flight altitudes (150, 200, 250, and 300\,m), while IRVL328~\cite{liu2025object} considers infrared-to-visible retrieval. No fine-tuning or target-side model selection is performed on SUES-200 or IRVL328.\\
\textbf{Baselines.}
We compare OCL with both satellite-accessed DVGL methods and target-satellite-free place-recognition and geo-localization baselines.
The former group includes Instance Loss~\cite{zheng2020dual}, FSRA~\cite{dai2021transformer}, Sample4Geo~\cite{deuser2023sample4geo}, Safe-Net~\cite{lin2024self}, MFRGN~\cite{wang2024mfrgn}, WeatherPrompt~\cite{wen2025WeatherPrompt}, SpikeViMFormer~\cite{chen2026efficient}, EM-CVGL~\cite{li2024learning}, UCVGL~\cite{li2024unleashing}, and DMNIL~\cite{chen2026without}.
For the target-satellite-free setting, we evaluate EigenPlaces~\cite{berton2023eigenplaces}, AnyLoc~\cite{keetha2023anyloc}, SALAD~\cite{izquierdo2024optimal}, and the training-free cross-view method VFM-Loc~\cite{lu2026vfmloc}.
We additionally construct an architecture-matched source-UAV baseline, denoted as \emph{Ours (NCE-only)}.
It shares the DINOv3 backbone, ViewTokenizer, OrbitFormer, observation curriculum, University-1652 training data, and optimization budget with OCL, but uses only real UAV observations and scene-prototype InfoNCE, without rendered views or completed-orbit distillation.
Both Ours (NCE-only) and OCL are trained and selected exclusively on University-1652 and then evaluated with frozen checkpoints on SUES-200 and IRVL328.\\
\textbf{Evaluation metrics.}
In the main paper, we report Recall@1 (R@1) and mean Average Precision (mAP) for drone-to-satellite retrieval. Further R@5, R@10, and MRR results are provided in the supplementary material. For SUES-200, results are reported separately at 150, 200, 250, and 300\,m, while IRVL328 evaluates infrared UAV queries against a visible satellite gallery.\\
\textbf{Implementation details.}
We use DINOv3 ViT-B/16~\cite{simeoni2025dinov3} as the image backbone with $256\times256$ inputs.
ViewTokenizer uses four learned queries, and OrbitFormer contains nine latent tokens, producing a 768-D L2-normalized descriptor.
Models are trained for 20 epochs (2000 optimization steps) with AdamW~\cite{loshchilov2017decoupled}.
The learning rates are $10^{-6}$ for the backbone and $10^{-4}$ for newly introduced modules, with weight decay 0.05, layer-wise learning-rate decay 0.8, and a 5\% warmup followed by cosine decay.
We use mixed-precision training, gradient clipping at 1.0, and a logical batch size of eight scenes.
Unless otherwise stated, controlled experiments use seed 0.
Further implementation details are provided in the supplementary material.

\subsection{Main Benchmark Results}
\label{sec:main_results}

\begin{table*}[t]
\centering
\small
\renewcommand{\arraystretch}{1.05}

\resizebox{\textwidth}{!}{%
\begin{tabular}{@{}llcccccccccccc@{}}
\toprule
& &
\multicolumn{2}{c}{University-1652} &
\multicolumn{8}{c}{SUES-200} &
\multicolumn{2}{c}{IRVL328} \\
\cmidrule(lr){3-4}
\cmidrule(lr){5-12}
\cmidrule(lr){13-14}

Method & Pub. &
\multicolumn{2}{c}{Drone$\rightarrow$Sat.} &
\multicolumn{2}{c}{150\,m} &
\multicolumn{2}{c}{200\,m} &
\multicolumn{2}{c}{250\,m} &
\multicolumn{2}{c}{300\,m} &
\multicolumn{2}{c}{IR$\rightarrow$Vis.} \\

& &
R@1 & mAP &
R@1 & mAP &
R@1 & mAP &
R@1 & mAP &
R@1 & mAP &
R@1 & mAP \\
\midrule

\multicolumn{14}{@{}l}{\textit{Satellite-accessed before final retrieval}} \\

Instance Loss~\cite{zheng2020dual} & MM'20
& 56.89 & 66.41
& 23.77 & 38.12
& 30.55 & 43.30
& 31.95 & 44.32
& 33.90 & 45.26
& 4.66 & 10.66 \\

FSRA~\cite{dai2021transformer} & TCSVT'21
& \best{98.65} & \best{99.29}
& 51.73 & 64.80
& 61.58 & 73.33
& 68.00 & 78.30
& 72.72 & 81.61
& 11.63 & 21.38 \\

Sample4Geo~\cite{deuser2023sample4geo} & ICCV'23
& 92.11 & 94.57
& 70.58 & \second{79.12}
& \second{81.53} & \second{87.71}
& \second{86.58} & \second{91.25}
& \second{90.12} & \second{93.60}
& 16.53 & 28.29 \\

Safe-Net~\cite{lin2024self} & TIP'24
& 80.42 & 85.89
& 60.98 & 71.82
& 71.43 & 80.04
& 79.17 & 85.49
& 81.60 & 87.23
& \second{23.95} & \second{34.78} \\

MFRGN~\cite{wang2024mfrgn} & MM'24
& \second{93.55} & \second{95.44}
& 66.17 & 75.59
& 78.05 & 84.67
& 85.55 & 89.75
& 87.55 & 91.15
& 18.95 & 28.52 \\

WeatherPrompt~\cite{wen2025WeatherPrompt} & NeurIPS'25
& 82.40 & 87.27
& 53.42 & 64.91
& 65.67 & 75.86
& 74.02 & 82.73
& 80.65 & 87.47
& 15.24 & 26.22 \\

SpikeViMFormer~\cite{chen2026efficient} & TCSVT'26
& 87.91 & 91.89
& \second{70.60} & 78.91
& 80.20 & 86.22
& 84.75 & 89.26
& 87.15 & 90.99
& 18.34 & 29.02 \\

EM-CVGL~\cite{li2024learning} & TGRS'24
& 70.22 & 79.42
& 61.50 & 72.46
& 72.80 & 81.87
& 81.67 & 88.12
& 86.78 & 91.44
& 10.01 & 21.49 \\

UCVGL~\cite{li2024unleashing} & CVPR'24
& 36.59 & 44.89
& 21.10 & 31.38
& 30.03 & 40.56
& 35.05 & 45.50
& 38.75 & 49.41
& 9.15 & 15.82 \\

DMNIL~\cite{chen2026without} & TNNLS'26
& 89.35 & 92.61
& \best{76.95} & \best{84.41}
& \best{86.52} & \best{91.41}
& \best{91.12} & \best{94.54}
& \best{93.97} & \best{96.47}
& \best{35.40} & \best{49.78} \\

\midrule

\multicolumn{14}{@{}l}{\textit{Target-satellite-free}} \\

EigenPlaces~\cite{berton2023eigenplaces} & ICCV'23
& 4.60 & 7.86
& 12.15 & 20.03
& 15.40 & 24.01
& 18.52 & 27.17
& 21.02 & 29.01
& 3.03 & 6.90 \\

AnyLoc~\cite{keetha2023anyloc} & RA-L'23
& 17.08 & 25.49
& 34.83 & 48.09
& 40.98 & 53.46
& 46.75 & 58.64
& 48.63 & 60.77
& \second{5.33} & 13.20 \\

SALAD~\cite{izquierdo2024optimal} & CVPR'24
& 8.43 & 14.34
& 15.57 & 24.12
& 21.62 & 30.30
& 25.55 & 34.93
& 30.50 & 40.13
& 3.86 & 9.09 \\

VFM-Loc$^\dagger$~\cite{lu2026vfmloc} & arXiv'26
& 33.31 & 44.90
& 35.15 & 48.47
& 43.95 & 58.33
& 54.90 & \second{67.48}
& \best{68.17} & \best{77.70}
& 4.97 & 12.75 \\

Ours (NCE-only) & --
& \second{49.62} & \second{61.05}
& \second{36.88} & \second{49.38}
& \second{48.33} & \second{59.83}
& \second{55.33} & 66.06
& 63.78 & 72.74
& 5.16 & \second{13.37} \\

\textbf{Ours (OCL)} & --
& \best{57.85} & \best{68.44}
& \best{46.48} & \best{57.23}
& \best{54.48} & \best{65.01}
& \best{60.30} & \best{69.81}
& \second{66.33} & \second{74.60}
& \best{5.95} & \best{15.11} \\

\bottomrule
\end{tabular}%
}

\caption{
Cross-benchmark drone-to-satellite retrieval (\%).
All methods are trained on University-1652 and evaluated on University-1652, SUES-200, and IRVL328.
Methods are grouped by satellite access during task-specific development.
\textcolor{red}{\textbf{Red}} and \textcolor{blue}{\textbf{blue}} indicate the best and second-best results within each protocol group.
$^\dagger$VFM-Loc uses UAV-query-only PCA in our evaluation.
}
\label{tab:main_benchmarks}
\end{table*}
Table~\ref{tab:main_benchmarks} summarizes the cross-benchmark retrieval results.
\textbf{Results on University-1652.}
OCL achieves 57.85\% R@1 and 68.44\% mAP, ranking first among all target-satellite-free methods. It also outperforms the satellite-accessed Instance Loss baseline. Compared with the architecture-matched NCE-only baseline, OCL improves R@1 and mAP by 8.23 and 7.39 points, respectively, demonstrating the effectiveness of orbit completion.\\
\textbf{Results on SUES-200.}
Under direct cross-dataset transfer, OCL achieves the best target-satellite-free results at 150\,m, 200\,m, and 250\,m, and ranks second at 300\,m. Compared with NCE-only, the R@1 gains are 9.60, 6.15, 4.97, and 2.55 points from 150\,m to 300\,m. This trend indicates that orbit completion is particularly beneficial under larger viewpoint gaps.\\
\textbf{Results on IRVL328.}
OCL achieves the best target-satellite-free performance with 5.95\% R@1 and 15.11\% mAP, outperforming NCE-only by 0.79 and 1.74 points, respectively. The gain under direct transfer further shows that orbit completion improves representation transferability across a substantial modality gap.

\subsection{Robustness to Weather-Style Appearance Shifts}
\label{sec:weather}

\begin{table*}[t]
\centering
\small
\renewcommand{\arraystretch}{1.05}

\resizebox{\textwidth}{!}{%
\begin{tabular}{@{}lcccccccccccccccccc@{}}
\toprule
Method
& \multicolumn{2}{c}{Fog}
& \multicolumn{2}{c}{Rain}
& \multicolumn{2}{c}{Snow}
& \multicolumn{2}{c}{Fog+Rain}
& \multicolumn{2}{c}{Fog+Snow}
& \multicolumn{2}{c}{Rain+Snow}
& \multicolumn{2}{c}{Dark}
& \multicolumn{2}{c}{Over-exp.}
& \multicolumn{2}{c}{Wind} \\
\cmidrule(lr){2-3}
\cmidrule(lr){4-5}
\cmidrule(lr){6-7}
\cmidrule(lr){8-9}
\cmidrule(lr){10-11}
\cmidrule(lr){12-13}
\cmidrule(lr){14-15}
\cmidrule(lr){16-17}
\cmidrule(lr){18-19}

& R@1 & mAP
& R@1 & mAP
& R@1 & mAP
& R@1 & mAP
& R@1 & mAP
& R@1 & mAP
& R@1 & mAP
& R@1 & mAP
& R@1 & mAP \\
\midrule

\multicolumn{19}{@{}l}{\textit{Satellite-accessed before final retrieval}} \\

Instance Loss~\cite{zheng2020dual}
& 0.29 & 2.03
& 1.13 & 3.34
& 1.68 & 4.13
& 0.05 & 0.74
& 0.18 & 0.95
& 1.00 & 2.82
& 0.01 & 0.62
& 4.95 & 10.20
& 13.84 & 22.43 \\

FSRA~\cite{dai2021transformer}
& 31.46 & 42.04
& 24.03 & 31.55
& 15.45 & 24.24
& 3.36 & 6.47
& 2.20 & 4.91
& 18.97 & 27.39
& 2.43 & 5.02
& 63.86 & 72.88
& \best{91.62} & \best{94.85} \\

Sample4Geo~\cite{deuser2023sample4geo}
& \second{87.39} & \second{91.48}
& 53.62 & 62.20
& 25.99 & 37.08
& 44.68 & 53.80
& 7.88 & 14.86
& 24.06 & 34.16
& 33.28 & 40.77
& 65.42 & 73.90
& 82.08 & 87.45 \\

Safe-Net~\cite{lin2024self}
& 35.76 & 46.15
& 23.12 & 30.92
& 13.92 & 21.05
& 10.58 & 16.83
& 2.86 & 6.01
& 15.67 & 22.90
& 13.11 & 19.60
& 43.33 & 53.40
& 63.32 & 71.65 \\

MFRGN~\cite{wang2024mfrgn}
& \best{88.17} & \best{91.79}
& 50.39 & 58.73
& 35.11 & 45.16
& 45.72 & 54.59
& 19.85 & 27.72
& \second{35.09} & 44.41
& 23.80 & 30.03
& 70.88 & 78.03
& \second{86.05} & \second{90.26} \\

WeatherPrompt~\cite{wen2025WeatherPrompt}
& 80.75 & 86.05
& \best{79.90} & \best{85.40}
& \best{76.45} & \best{82.72}
& \best{78.59} & \best{84.38}
& \best{72.42} & \best{79.47}
& \best{77.71} & \best{83.69}
& \best{68.69} & \best{76.03}
& \second{73.58} & \second{80.51}
& 75.78 & 82.22 \\

SpikeViMFormer~\cite{chen2026efficient}
& 69.27 & 77.53
& 37.57 & 46.82
& 13.70 & 22.45
& 23.14 & 31.59
& 3.42 & 6.72
& 16.87 & 24.98
& 21.50 & 28.69
& 65.46 & 74.51
& 60.34 & 69.50 \\

EM-CVGL~\cite{li2024learning}
& 60.91 & 71.65
& 47.53 & 59.04
& \second{38.70} & \second{49.74}
& 39.61 & 51.19
& \second{24.56} & \second{34.80}
& 33.88 & \second{44.52}
& \second{44.19} & \second{54.81}
& 57.51 & 68.74
& 58.49 & 69.39 \\

UCVGL~\cite{li2024unleashing}
& 32.67 & 40.70
& 17.17 & 23.23
& 4.88 & 9.07
& 15.67 & 21.25
& 1.74 & 4.68
& 8.42 & 13.16
& 20.82 & 27.24
& 29.62 & 37.60
& 13.85 & 20.04 \\

DMNIL~\cite{chen2026without}
& 85.90 & 90.12
& \second{57.77} & \second{65.78}
& 21.87 & 30.38
& \second{55.25} & \second{63.25}
& 8.65 & 13.95
& 31.74 & 40.87
& 42.01 & 49.29
& \best{76.56} & \best{82.79}
& 52.88 & 61.60 \\

\midrule

\multicolumn{19}{@{}l}{\textit{Target-satellite-free}} \\

EigenPlaces~\cite{berton2023eigenplaces}
& 3.85 & 6.91
& 1.77 & 3.97
& 2.74 & 5.22
& 2.11 & 4.57
& 2.37 & 4.71
& 1.85 & 4.12
& 3.12 & 5.81
& 3.33 & 6.24
& 0.71 & 2.00 \\

AnyLoc~\cite{keetha2023anyloc}
& 28.98 & 39.93
& 16.89 & 26.47
& 20.08 & 29.76
& 15.10 & 24.41
& \best{13.99} & \best{22.47}
& 14.29 & 22.99
& 16.38 & 24.14
& 24.34 & 34.63
& 22.44 & 32.66 \\

SALAD~\cite{izquierdo2024optimal}
& 8.60 & 14.76
& 4.85 & 9.34
& 6.07 & 11.33
& 5.16 & 9.91
& 5.42 & 10.22
& 4.48 & 8.91
& 5.19 & 9.94
& 5.98 & 11.19
& 6.05 & 11.30 \\

VFM-Loc$^\dagger$~\cite{lu2026vfmloc}
& 36.82 & 48.49
& \second{25.06} & \second{36.18}
& \second{20.39} & \second{30.67}
& \second{20.56} & \second{31.33}
& \second{13.88} & \second{22.40}
& \second{19.61} & \best{29.76}
& \second{23.33} & \second{33.10}
& \second{32.79} & \second{44.33}
& 36.17 & 47.72 \\

Ours (NCE-only)
& \second{36.95} & \second{48.52}
& 24.48 & 34.79
& 19.21 & 28.69
& 18.81 & 28.37
& 11.71 & 19.08
& 16.59 & 25.31
& 22.78 & 32.05
& 30.29 & 41.24
& \second{37.18} & \second{48.59} \\

\textbf{Ours (OCL)}
& \best{44.32} & \best{55.91}
& \best{29.21} & \best{39.71}
& \best{22.27} & \best{32.29}
& \best{23.02} & \best{33.02}
& 13.41 & 21.46
& \best{20.20} & \second{29.59}
& \best{26.92} & \best{36.40}
& \best{38.90} & \best{50.01}
& \best{43.87} & \best{55.12} \\

\bottomrule
\end{tabular}%
}

\caption{
Robustness to weather-style appearance shifts on University-1652.
Weather transformations are applied only to UAV queries, while the satellite gallery remains unchanged.
\textcolor{red}{\textbf{Red}} and \textcolor{blue}{\textbf{blue}} indicate the best and second-best results within each protocol group.
$^\dagger$VFM-Loc uses UAV-query-only PCA in our evaluation.
}
\label{tab:weather}
\end{table*}

\begin{figure}[t]
    \centering
    \includegraphics[width=\columnwidth]{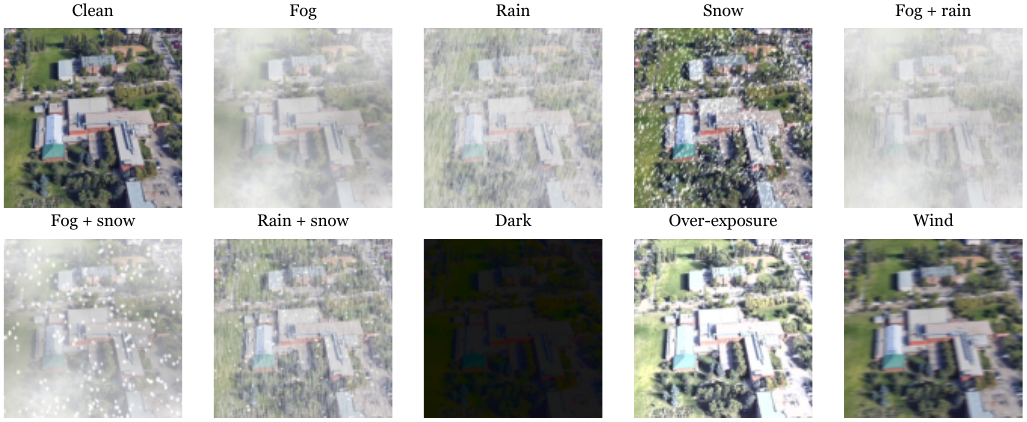}
    \caption{Examples of the weather-style appearance shifts used for robustness evaluation on University-1652. Starting from a clean UAV query, we synthesize nine challenging conditions, including fog, rain, snow, their compound combinations, darkness, over-exposure, and wind-induced motion blur.}
    \label{fig:weather_transforms}
\end{figure}
We further evaluate robustness to weather-style appearance shifts on University-1652, following~\cite{wang2024multiple,wen2025WeatherPrompt}. Nine transformations are applied only to UAV queries, including fog, rain, snow, their compound conditions, darkness, over-exposure, and wind-induced motion blur, while the satellite gallery remains unchanged. Representative transformations are shown in Fig.~\ref{fig:weather_transforms}, and the quantitative results are reported in Table~\ref{tab:weather}. Implementation details are provided in the supplementary material.\\
\textbf{Generalization to unseen weather conditions.}
Without weather-specific training, OCL achieves the best R@1 in eight of nine conditions and the best mAP in seven among target-satellite-free methods. More notably, despite using no target satellite imagery during development, OCL outperforms several satellite-accessed methods under multiple shifts, surpassing both FSRA and Safe-Net in R@1 and mAP across seven of the nine conditions. OCL also improves over the architecture-matched NCE-only baseline under every condition, showing that orbit completion enhances generalization beyond source-only adaptation. The remaining weakness under severe compound shifts, particularly fog+snow, indicates that extreme appearance corruption remains challenging.

\subsection{Qualitative Analysis}
\label{sec:qualitative}

\begin{figure*}[t]
    \centering
    \includegraphics[width=\textwidth]{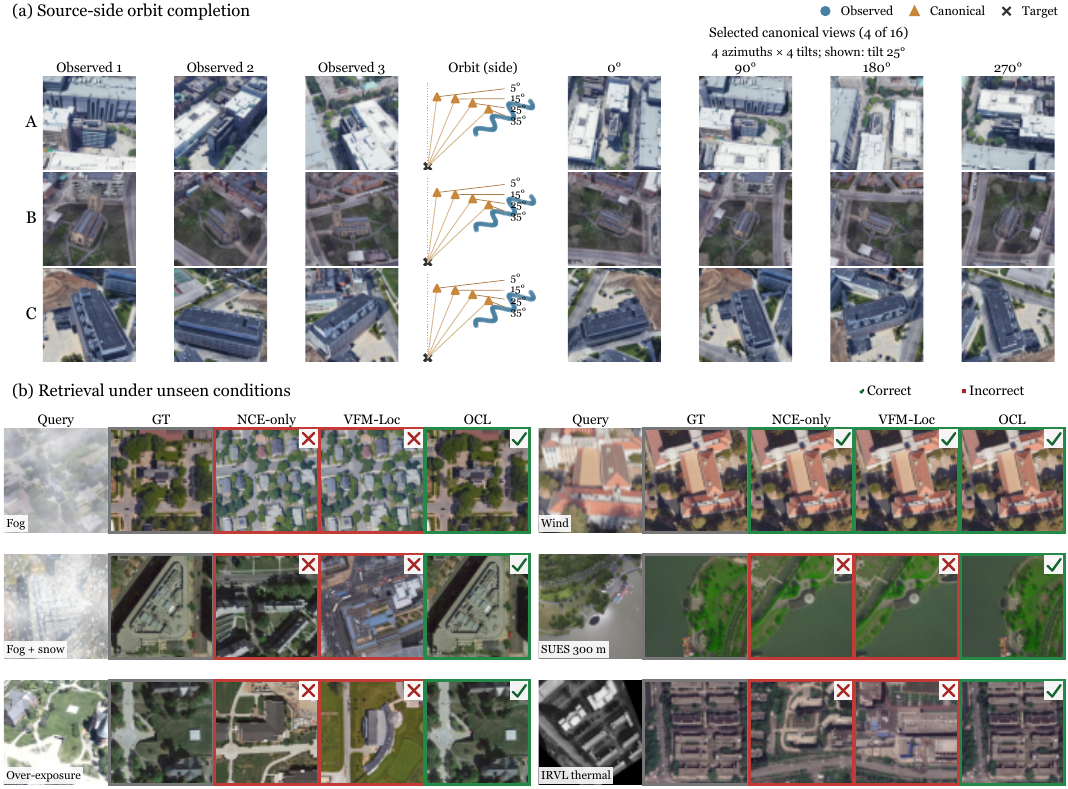}
    \caption{
    Qualitative analysis of OCL.
    (a) Source-side orbit completion. OCL reconstructs each source scene from observed UAV images and renders a 16-view canonical orbit covering four azimuths and four tilts; four views at $25^\circ$ tilt are shown.
    (b) Retrieval under unseen appearance, viewpoint, and modality shifts.
    Green and red borders indicate correct and incorrect retrievals, respectively.
    OCL is developed without target satellite imagery or weather-specific supervision.
    }
    \label{fig:qualitative}
\end{figure*}

Fig.~\ref{fig:qualitative}(a) illustrates how OCL expands limited source observations into a structured canonical orbit. \textbf{Despite substantial viewpoint changes, the rendered views preserve the dominant scene layout and building geometry across azimuths and tilts.} This provides complementary observations of the same place and encourages the learned representation to focus on scene-level structure rather than a particular UAV appearance or viewing direction. The retrieval examples in Fig.~\ref{fig:qualitative}(b) further show that \textbf{this structural supervision transfers across appearance, viewpoint, and modality shifts.} Under fog, fog+snow, and over-exposure, NCE-only and VFM-Loc are often distracted by candidates with similar coarse appearance but incorrect spatial layouts, whereas OCL retrieves the correct location. Similar behavior is observed under the large viewpoint change of SUES-200 at 300\,m and the infrared-to-visible gap on IRVL328. Under wind-induced blur, all three methods succeed when the underlying scene structure remains sufficiently recognizable. Additional success and failure cases are provided in the supplementary material.

\subsection{Ablation Study}
\label{sec:ablation}

\begin{table}[t]
\centering
\small
\setlength{\tabcolsep}{3.2pt}
\renewcommand{\arraystretch}{1.06}

\begin{tabular}{@{}lcccccc@{}}
\toprule
Setting
& Student
& Teacher
& Dist.
& NCE
& R@1
& mAP \\
\midrule

\multicolumn{7}{@{}l}{\textit{(a) Student and teacher completion}} \\

Real only
& R & R
& \checkmark & \checkmark
& 51.17 & 62.55 \\

Teacher completion
& R & R+C
& \checkmark & \checkmark
& 53.24 & 64.39 \\

Student augmentation
& R+C & R
& \checkmark & \checkmark
& 55.71 & 66.57 \\

\textbf{Full OCL}
& \textbf{R+C} & \textbf{R+C}
& \checkmark & \checkmark
& \textbf{57.85} & \textbf{68.44} \\

\midrule

\multicolumn{7}{@{}l}{\textit{(b) Geometry versus appearance}} \\

Real context
& R & R
& \checkmark & \checkmark
& 51.17 & 62.55 \\

+ Appearance
& R & R+A
& \checkmark & \checkmark
& 50.51 & 61.96 \\

+ Canonical
& R & R+C
& \checkmark & \checkmark
& \textbf{53.24} & \textbf{64.39} \\

\midrule

\multicolumn{7}{@{}l}{\textit{(c) Training objectives}} \\

\textbf{Full OCL}
& \textbf{R+C} & \textbf{R+C}
& \checkmark & \checkmark
& \textbf{57.85} & \textbf{68.44} \\

w/o Distillation
& R+C & R+C
& -- & \checkmark
& 55.18 & 65.88 \\

w/o InfoNCE
& R+C & R+C
& \checkmark & --
& 36.45 & 48.14 \\

\bottomrule
\end{tabular}

\caption{Ablation studies on University-1652 (\%).
R, C, and A denote real UAV observations, canonical 3DGS-rendered observations, and appearance-augmented observations, respectively.
All configurations follow the same training and model-selection protocol, with only the factors indicated in each comparison changed.
(a) separates student-side augmentation from teacher-side orbit completion;
(b) compares geometry-derived canonical views with number-matched appearance augmentation;
and (c) examines the contributions of the two training objectives.}
\label{tab:ablation_main}
\end{table}

We conduct controlled ablations on University-1652 to examine three aspects of OCL: the roles of canonical observations on the student and teacher sides, the contribution of geometric viewpoint coverage relative to appearance augmentation, and the two training objectives. All configurations use the same architecture, source split, optimization schedule, observation-count curriculum, and checkpoint-selection protocol, while varying only the component under investigation. Results are summarized in Table~\ref{tab:ablation_main}.\\
\textbf{Student augmentation and teacher completion.}
Table~\ref{tab:ablation_main}(a) shows that canonical observations benefit both sides of OCL. Adding canonical views only to the teacher improves R@1/mAP from 51.17/62.55 to 53.24/64.39, while student-side augmentation alone reaches 55.71/66.57. Combining both yields 57.85/68.44, improving over the real-only configuration by 6.68 R@1 and 5.89 mAP points. Notably, adding completed-orbit teacher supervision on top of student-side augmentation further improves R@1 by 2.14 points, showing that the teacher contributes beyond using rendered views merely as additional student training samples.\\
\textbf{Geometry versus appearance augmentation.}
Table~\ref{tab:ablation_main}(b) compares two number-matched extensions of the teacher context while keeping the student restricted to real UAV observations. Appearance augmentation does not improve the real-only baseline, yielding 50.51/61.96 compared with 51.17/62.55. In contrast, replacing the same number of appearance-augmented observations with canonical 3DGS views increases performance to 53.24/64.39, outperforming the appearance control by 2.73 R@1 and 2.43 mAP points. These results indicate that the benefit of teacher completion is associated with complementary geometric viewpoint coverage rather than simply increasing the number or appearance diversity of teacher observations.\\
\textbf{Training objectives.}
Table~\ref{tab:ablation_main}(c) shows that the two objectives play complementary roles. Removing distillation reduces R@1/mAP from 57.85/68.44 to 55.18/65.88, corresponding to drops of 2.67 and 2.56 points, respectively. Removing InfoNCE causes a substantially larger degradation to 36.45/48.14, with drops of 21.40 R@1 and 20.30 mAP points. This indicates that scene-level contrastive discrimination is essential for retrieval, while partial-to-completed distillation provides an additional benefit by transferring completed-orbit context to the student. 
Further ablation, variable-cardinality, rendering, and computational analyses are provided in the supplementary material.

\section{Conclusion}
We propose Orbit Completion Learning (OCL) for drone-view geo-localization with late-arriving satellite galleries, where target satellite imagery is unavailable during task-specific development. OCL uses source-side multi-view geometry as privileged supervision: canonical viewpoints are reconstructed to complete the observed UAV orbit, and the resulting scene context is distilled into a variable-cardinality encoder that operates on a single image at inference. Experiments on University-1652, SUES-200, IRVL328, and nine unseen weather-style shifts show consistent improvements in cross-view and cross-condition transfer without target-satellite supervision. Controlled ablations further indicate that the gains arise from complementary geometric viewpoints and completed-orbit supervision rather than synthetic appearance diversity alone. More broadly, when source-side geometry can provide meaningful surrogate viewpoints during training, missing target views need not be explicitly supplied from the target domain. Geometry is not for deployment, but for shaping transferable invariance during training. OCL turns multi-view geometry into an image representation, leaving inference as simple single-image retrieval.

\FloatBarrier

{\small
\bibliographystyle{ieeenat_fullname}
\bibliography{references}

@inproceedings{wen2025WeatherPrompt,
  author = "Wen, Jiahao and Yu, Hang and Zheng, Zhedong",
  title = "WeatherPrompt: Multi-modality Representation Learning for All-Weather Drone Visual Geo-Localization",
  booktitle = "NeurIPS",
  year = "2025" 
}

@article{wang2021each,
  title={Each part matters: Local patterns facilitate cross-view geo-localization},
  author={Wang, Tingyu and Zheng, Zhedong and Yan, Chenggang and Zhang, Jiyong and Sun, Yaoqi and Zheng, Bolun and Yang, Yi},
  journal={IEEE Transactions on Circuits and Systems for Video Technology},
  volume={32},
  number={2},
  pages={867--879},
  year={2021},
  publisher={IEEE}
}

@article{dai2021transformer,
  title={A transformer-based feature segmentation and region alignment method for UAV-view geo-localization},
  author={Dai, Ming and Hu, Jianhong and Zhuang, Jiedong and Zheng, Enhui},
  journal={IEEE Transactions on Circuits and Systems for Video Technology},
  volume={32},
  number={7},
  pages={4376--4389},
  year={2021},
  publisher={IEEE}
}

@inproceedings{deuser2023sample4geo,
  title={Sample4geo: Hard negative sampling for cross-view geo-localisation},
  author={Deuser, Fabian and Habel, Konrad and Oswald, Norbert},
  booktitle={Proceedings of the IEEE/CVF International Conference on Computer Vision},
  pages={16847--16856},
  year={2023}
}

@article{keetha2023anyloc,
  title={Anyloc: Towards universal visual place recognition},
  author={Keetha, Nikhil and Mishra, Avneesh and Karhade, Jay and Jatavallabhula, Krishna Murthy and Scherer, Sebastian and Krishna, Madhava and Garg, Sourav},
  journal={IEEE Robotics and Automation Letters},
  volume={9},
  number={2},
  pages={1286--1293},
  year={2023},
  publisher={IEEE}
}

@article{li2024learning,
  title={Learning cross-view visual geo-localization without ground truth},
  author={Li, Haoyuan and Xu, Chang and Yang, Wen and Yu, Huai and Xia, Gui-Song},
  journal={IEEE Transactions on Geoscience and Remote Sensing},
  year={2024},
  publisher={IEEE}
}

@inproceedings{li2024unleashing,
  title={Unleashing unlabeled data: A paradigm for cross-view geo-localization},
  author={Li, Guopeng and Qian, Ming and Xia, Gui-Song},
  booktitle={Proceedings of the IEEE/CVF Conference on Computer Vision and Pattern Recognition},
  pages={16719--16729},
  year={2024}
}

@article{chen2025limited,
  title={From Limited Labels to Open Domains: An Efficient Learning Paradigm for UAV-view Geo-Localization},
  author={Chen, Zhongwei and Yang, Zhao-Xu and Rong, Hai-Jun and Lang, Jiawei},
  journal={arXiv preprint arXiv:2503.07520},
  year={2025}
}

@article{li2025unsupervised,
  title={Unsupervised Multi-view UAV Image Geo-localization via Iterative Rendering},
  author={Li, Haoyuan and Xu, Chang and Yang, Wen and Mi, Li and Yu, Huai and Zhang, Haijian and Xia, Gui-Song},
  journal={IEEE Transactions on Geoscience and Remote Sensing},
  year={2025},
  publisher={IEEE}
}

@InProceedings{Ju_2025_ICCV,
    author    = {Ju, Hao and Huang, Shaofei and Liu, Si and Zheng, Zhedong},
    title     = {Video2BEV: Transforming Drone Videos to BEVs for Video-based Geo-localization},
    booktitle = {Proceedings of the IEEE/CVF International Conference on Computer Vision (ICCV)},
    month     = {October},
    year      = {2025},
    pages     = {27073-27083}
}

@Article{kerbl3Dgaussians,
      author       = {Kerbl, Bernhard and Kopanas, Georgios and Leimk{\"u}hler, Thomas and Drettakis, George},
      title        = {3D Gaussian Splatting for Real-Time Radiance Field Rendering},
      journal      = {ACM Transactions on Graphics},
      number       = {4},
      volume       = {42},
      month        = {July},
      year         = {2023},
      url          = {https://repo-sam.inria.fr/fungraph/3d-gaussian-splatting/}
}

@inproceedings{zheng2020university,
  title={University-1652: A multi-view multi-source benchmark for drone-based geo-localization},
  author={Zheng, Zhedong and Wei, Yunchao and Yang, Yi},
  booktitle={Proceedings of the 28th ACM international conference on Multimedia},
  pages={1395--1403},
  year={2020}
}

@article{zhu2023sues,
  title={SUES-200: A multi-height multi-scene cross-view image benchmark across drone and satellite},
  author={Zhu, Runzhe and Yin, Ling and Yang, Mingze and Wu, Fei and Yang, Yuncheng and Hu, Wenbo},
  journal={IEEE Transactions on Circuits and Systems for Video Technology},
  volume={33},
  number={9},
  pages={4825--4839},
  year={2023},
  publisher={IEEE}
}

@article{zheng2020dual,
  title={Dual-path convolutional image-text embeddings with instance loss},
  author={Zheng, Zhedong and Zheng, Liang and Garrett, Michael and Yang, Yi and Xu, Mingliang and Shen, Yi-Dong},
  journal={ACM Transactions on Multimedia Computing, Communications, and Applications (TOMM)},
  volume={16},
  number={2},
  pages={1--23},
  year={2020},
  publisher={ACM New York, NY, USA}
}

@article{lin2024self,
  title={A self-adaptive feature extraction method for aerial-view geo-localization},
  author={Lin, Jinliang and Luo, Zhiming and Lin, Dazhen and Li, Shaozi and Zhong, Zhun},
  journal={IEEE Transactions on Image Processing},
  volume={34},
  pages={126--139},
  year={2024},
  publisher={IEEE}
}

@inproceedings{wang2024mfrgn,
  title={Mfrgn: Multi-scale feature representation generalization network for ground-to-aerial geo-localization},
  author={Wang, Yuntao and Zhang, Jinpu and Wei, Ruonan and Gao, Wenbo and Wang, Yuehuan},
  booktitle={Proceedings of the 32nd ACM International Conference on Multimedia},
  pages={2574--2583},
  year={2024}
}

@article{liu2026diffusionuavloc,
  title={DiffusionUavLoc: Visually Prompted Diffusion for Cross-View UAV Localization},
  author={Liu, Tao and Ren, Kan and Chen, Qian},
  journal={IEEE Internet of Things Journal},
  year={2026},
  publisher={IEEE}
}

@article{wang2025coarse, title={From Coarse to Fine: A Matching and Alignment Framework for Unsupervised Cross-View Geo-Localization}, volume={39}, url={https://ojs.aaai.org/index.php/AAAI/article/view/32865}, DOI={10.1609/aaai.v39i8.32865}, abstractNote={Cross-view geo-localization aims at determining the geographic location of a query image by matching the reference images. The matching pairs can be captured from diverse perspectives, such as those from satellites and drones. Most existing methods are supervised that require input of location-labeled images or matched and unmatched image pairs for training, resulting in high labor costs. Moreover, current unsupervised methods perform instances matching directly between different perspectives with dramatic discrepancies, resulting in poor performance. To address these issues, this paper proposes a novel matching and alignment framework from coarse instance-cluster level to fine intermediate instance level for unsupervised cross-view geo-localization. We first introduces cluster-based contrastive learning, assigning pseudo-labels to the instances and generate clusters within each view. Then we design a cross-view location alignment module that fully exploits the feature relationships between instances and clusters for intra- and inter-views. Finally, we design an intermediate state transition module that facilitates further alignment between views by constructing intermediate states and bringing both views closer to the intermediate domain simultaneously. Extensive experiments demonstrate that our method surpasses state-of-the-art unsupervised cross-view geo-localization methods and even achieves comparable performance to state-of-the-art supervised methods.}, number={8}, journal={Proceedings of the AAAI Conference on Artificial Intelligence}, author={Wang, Xueyi and Zhang, Lele and Fan, Zheng and Liu, Yang and Chen, Chen and Deng, Fang}, year={2025}, month={Apr.}, pages={8024–8032} }

@article{chen2026without,
  title={Without paired labeled data: End-to-end self-supervised learning for drone-view geo-localization},
  author={Chen, Zhongwei and Yang, Zhao-Xu and Rong, Hai-Jun and Li, Guoqi},
  journal={IEEE Transactions on Neural Networks and Learning Systems},
  year={2026},
  publisher={IEEE}
}

@inproceedings{arandjelovic2016netvlad,
  title={NetVLAD: CNN architecture for weakly supervised place recognition},
  author={Arandjelovic, Relja and Gronat, Petr and Torii, Akihiko and Pajdla, Tomas and Sivic, Josef},
  booktitle={Proceedings of the IEEE conference on computer vision and pattern recognition},
  pages={5297--5307},
  year={2016}
}

@article{li2025place,
  title={Place recognition meet multiple modalities: a comprehensive review, current challenges and future development},
  author={Li, Zhenyu and Shang, Tianyi and Xu, Pengjie and Deng, Zhaojun},
  journal={Artificial Intelligence Review},
  volume={58},
  number={11},
  pages={363},
  year={2025},
  publisher={Springer}
}

@inproceedings{ali2024boq,
  title={BoQ: A place is worth a bag of learnable queries},
  author={Ali-Bey, Amar and Chaib-Draa, Brahim and Giguere, Philippe},
  booktitle={2024 IEEE/CVF Conference on Computer Vision and Pattern Recognition (CVPR)},
  pages={17794--17803},
  year={2024},
  organization={IEEE}
}

@article{simeoni2025dinov3,
  title={Dinov3},
  author={Sim{\'e}oni, Oriane and Vo, Huy V and Seitzer, Maximilian and Baldassarre, Federico and Oquab, Maxime and Jose, Cijo and Khalidov, Vasil and Szafraniec, Marc and Yi, Seungeun and Ramamonjisoa, Micha{\"e}l and others},
  journal={arXiv preprint arXiv:2508.10104},
  year={2025}
}

@article{dai2023vision,
  title={Vision-based UAV self-positioning in low-altitude urban environments},
  author={Dai, Ming and Zheng, Enhui and Feng, Zhenhua and Qi, Lei and Zhuang, Jiedong and Yang, Wankou},
  journal={IEEE Transactions on Image Processing},
  volume={33},
  pages={493--508},
  year={2023},
  publisher={IEEE}
}

@article{liu2025object,
  title={Object Detection as an Optional Basis: A Graph Matching Network for Cross-View UAV Localization},
  author={Liu, Tao and Ren, Kan and Chen, Qian},
  journal={arXiv preprint arXiv:2511.02489},
  year={2025}
}

@article{lu2026vfmloc,
  title   = {VFM-Loc: Zero-Shot Cross-View Geo-Localization via Aligning Discriminative Visual Hierarchies},
  author  = {Jun Lu and Zehao Sang and Haoqi Wei and Xiangyun Liu and Kun Zhu and Haitao Guo and Zhihui Gong and Lei Ding},
  journal = {arXiv preprint arXiv:2603.13855},
  year    = {2026}
}

@inproceedings{schoenberger2016sfm,
  title={Structure-from-motion revisited},
  author={Schonberger, Johannes L and Frahm, Jan-Michael},
  booktitle={Proceedings of the IEEE conference on computer vision and pattern recognition},
  pages={4104--4113},
  year={2016}
}

@inproceedings{lee2019settransformer,
  title={Set Transformer: A Framework for Attention-based Permutation-Invariant Neural Networks},
  author={Lee, Juho and Lee, Yoonho and Kim, Jungtaek and Kosiorek, Adam and Choi, Seungjin and Teh, Yee Whye},
  booktitle={Proceedings of the 36th International Conference on Machine Learning},
  pages={3744--3753},
  year={2019}
}

@inproceedings{jaegle2021perceiver,
  title={Perceiver: General Perception with Iterative Attention},
  author={Jaegle, Andrew and Gimeno, Felix and Brock, Andrew and Vinyals, Oriol and Zisserman, Andrew and Carreira, Joao},
  booktitle={Proceedings of the 38th International Conference on Machine Learning},
  pages={4651--4664},
  year={2021}
}

@article{chen2026efficient,
  title={Efficient Spike-driven Transformer for High-performance Drone-View Geo-Localization},
  author={Chen, Zhongwei and Rong, Hai-Jun and Xu, Longrun and Lu, Jie and Zou, Zhuo and Yang, Zhao-Xu and Li, Guoqi},
  journal={IEEE Transactions on Circuits and Systems for Video Technology},
  year={2026},
  publisher={IEEE}
}

@inproceedings{berton2023eigenplaces,
  title={Eigenplaces: Training viewpoint robust models for visual place recognition},
  author={Berton, Gabriele and Trivigno, Gabriele and Caputo, Barbara and Masone, Carlo},
  booktitle={2023 IEEE/CVF International Conference on Computer Vision (ICCV)},
  pages={11046--11056},
  year={2023},
  organization={IEEE}
}

@inproceedings{izquierdo2024optimal,
  title={Optimal transport aggregation for visual place recognition},
  author={Izquierdo, Sergio and Civera, Javier},
  booktitle={2024 IEEE/CVF Conference on Computer Vision and Pattern Recognition (CVPR)},
  pages={17658--17668},
  year={2024},
  organization={IEEE}
}

@article{oquab2024dinov,
    title={{DINO}v2: Learning Robust Visual Features without Supervision},
    author={Maxime Oquab and Timoth{\'e}e Darcet and Th{\'e}o Moutakanni and Huy V. Vo and Marc Szafraniec and Vasil Khalidov and Pierre Fernandez and Daniel HAZIZA and Francisco Massa and Alaaeldin El-Nouby and Mido Assran and Nicolas Ballas and Wojciech Galuba and Russell Howes and Po-Yao Huang and Shang-Wen Li and Ishan Misra and Michael Rabbat and Vasu Sharma and Gabriel Synnaeve and Hu Xu and Herve Jegou and Julien Mairal and Patrick Labatut and Armand Joulin and Piotr Bojanowski},
    journal={Transactions on Machine Learning Research},
    issn={2835-8856},
    year={2024},
    url={https://openreview.net/forum?id=a68SUt6zFt},
    note={Featured Certification}
}

@article{loshchilov2017decoupled,
  title={Decoupled weight decay regularization},
  author={Loshchilov, Ilya and Hutter, Frank},
  journal={arXiv preprint arXiv:1711.05101},
  year={2017}
}

@article{wang2024multiple,
  title={Multiple-environment self-adaptive network for aerial-view geo-localization},
  author={Wang, Tingyu and Zheng, Zhedong and Sun, Yaoqi and Yan, Chenggang and Yang, Yi and Chua, Tat-Seng},
  journal={Pattern Recognition},
  volume={152},
  pages={110363},
  year={2024},
  publisher={Elsevier}
}
}

\appendix
\setcounter{figure}{0}
\renewcommand{\thefigure}{S\arabic{figure}}
\setcounter{table}{0}
\renewcommand{\thetable}{S\arabic{table}}
\setcounter{equation}{0}
\renewcommand{\theequation}{S\arabic{equation}}
\maketitlesupplementary



\section{Additional Implementation Details}
\label{sup:implementation}

\subsection{Model and Optimization}

We use DINOv3 ViT-B/16 as the image backbone with $256\times256$ inputs. ViewTokenizer contains four learned queries and two cross-attention blocks, each with eight attention heads and a 3072-D GELU feed-forward network. OrbitFormer contains one PLACE latent and eight auxiliary latents, and outputs a 768-D L2-normalized place descriptor. The student encoder contains 123.4M trainable parameters in total, including 85.6M in the DINOv3 backbone, 14.2M in ViewTokenizer, and 23.6M in OrbitFormer. During training, the EMA teacher mirrors the student architecture but is not optimized by gradient descent and is discarded after training.

Models are trained for 20 epochs (2000 optimization steps) with AdamW. The learning rates are $10^{-6}$ for the backbone and $10^{-4}$ for newly introduced modules, with weight decay 0.05 and layer-wise learning-rate decay 0.8. We use a 5\% linear warmup followed by cosine decay, mixed-precision training, gradient clipping at 1.0, a logical batch size of eight scenes, and a microbatch size of two scenes. Unless otherwise stated, controlled experiments use seed 0. All experiments are implemented with Python 3.9, PyTorch 2.8.0+cu128, CUDA 12.8, cuDNN 9.10.2, and timm 1.0.22, and are run on a single NVIDIA RTX 4090 D GPU.

\subsection{Observation-Count Curriculum}
During training, OCL progressively shifts the student from richer multi-view inputs toward the single-image operating point used at inference. The observation-count curriculum is shown in Table~\ref{tab:sup_curriculum}. For each training sample, the student draws from the observed-only pool with probability 0.7 and from the combined observed-and-rendered pool with probability 0.3. When $m>1$, at least one real UAV observation is included.

\begin{table}[t]
\centering
\small
\setlength{\tabcolsep}{4.0pt}
\begin{tabular}{lccccc}
\toprule
Progress & $m{=}16$ & $m{=}8$ & $m{=}4$ & $m{=}2$ & $m{=}1$\\
\midrule
0--25\%   & .40 & .35 & .15 & .05 & .05\\
25--60\%  & --  & .25 & .30 & .25 & .20\\
60--100\% & --  & --  & .15 & .30 & .55\\
\bottomrule
\end{tabular}
\caption{Observation-count curriculum used for OCL training. In the final 40\% of training, at least half of the scene samples in each logical batch are additionally forced to use a single observation.}
\label{tab:sup_curriculum}
\end{table}

\subsection{Source Split and Checkpoint Selection}
The 701 University-1652 source-side development scenes are natural-sorted, permuted using default\_rng(0), and divided into 631 optimization scenes and 70 validation scenes. Validation scenes are excluded from gradient-based representation learning, and all checkpoints are selected exclusively using this source-side validation split. Each training epoch consists of 100 optimization steps, yielding 2000 optimization steps over 20 epochs. For validation, singleton student descriptors are matched against scene-level teacher prototypes. Checkpoints are ranked lexicographically by mean R@1, mean MRR, mean hard-negative margin, and then earlier epoch, where the hard-negative margin is the cosine-similarity difference between the correct-scene prototype and the most similar incorrect-scene prototype.

\subsection{Token Memory and Prototype Bank}
The completed-orbit teacher maintains cached ViewTokenizer outputs for individual observations, $\Mmem_i\in\mathbb R^{|\Oset_i|\times4\times768}$. During optimization, sampled entries are updated as
\begin{equation}
\Mmem_{i,v}\leftarrow0.9\,\Mmem_{i,v}+0.1\,\bar{\bm V}_{i,v},
\end{equation}
where $\bar{\bm V}_{i,v}$ denotes the current EMA-teacher ViewTokenizer output. At the beginning of each epoch, a rotating 20\% subset of cached observations is re-encoded with the current EMA teacher. Cached observations are then independently processed through the epoch-start EMA OrbitFormer using the singleton path, and their normalized descriptors are averaged and normalized to form detached scene-level prototypes that remain fixed for the remainder of that epoch.

\section{Additional Geometry Details}
\label{sup:geometry}

\subsection{Reconstruction and Canonical Rendering}

Each source scene is reconstructed with COLMAP and subsequently optimized with 3D Gaussian Splatting. The key reconstruction and optimization settings are summarized in Table~\ref{tab:sup_geometry_config}. Canonical observations are rendered from the final 7000-step Gaussian snapshot. OCL then constructs four azimuths $\{0^\circ,90^\circ,180^\circ,270^\circ\}$ and four tilts $\{35^\circ,25^\circ,15^\circ,5^\circ\}$, yielding 16 complementary views per scene.

\begin{table}[t]
\centering
\small
\setlength{\tabcolsep}{4.2pt}
\renewcommand{\arraystretch}{1.05}
\begin{tabular}{ll}
\toprule
Setting & Value \\
\midrule
\multicolumn{2}{l}{\textit{COLMAP reconstruction}} \\
Pipeline & Sparse automatic reconstruction \\
Camera model & SIMPLE\_PINHOLE \\
Input resolution & $512\times512$ \\
Focal length & 614.4 px \\
Feature extraction & Covariant SIFT \\
Image matching & Exhaustive \\
Mapping & Incremental \\
Model selection & Most registered images \\
\midrule
\multicolumn{2}{l}{\textit{3D Gaussian Splatting}} \\
Optimization steps & 7000 \\
SH degree & 3 \\
Position LR & $1.6{\times}10^{-4}\rightarrow1.6{\times}10^{-6}$ \\
Position LR delay multiplier & 0.01 \\
DC feature LR & $2.5{\times}10^{-3}$ \\
Higher-order feature LR & $1.25{\times}10^{-4}$ \\
Opacity LR & $2.5{\times}10^{-2}$ \\
Scaling LR & $5{\times}10^{-3}$ \\
Rotation LR & $1{\times}10^{-3}$ \\
Photometric loss & $0.8\,L_1+0.2\,D\text{-SSIM}$ \\
Densification interval & 100 \\
Densification start & 500 \\
Densification threshold & $2{\times}10^{-4}$ \\
Percent dense & 0.01 \\
Opacity reset interval & 3000 \\
\midrule
\multicolumn{2}{l}{\textit{Canonical rendering}} \\
Azimuths & $0^\circ,90^\circ,180^\circ,270^\circ$ \\
Tilts & $35^\circ,25^\circ,15^\circ,5^\circ$ \\
Canonical views / scene & 16 \\
\bottomrule
\end{tabular}
\caption{
Reconstruction and canonical-rendering configuration used for OCL.
The 3DGS model is optimized for 7000 iterations and only the final snapshot is used to render canonical training observations.
}
\label{tab:sup_geometry_config}
\end{table}

\subsection{Robust Target Intersection}
Equation~\eqref{eq:target} is solved using at most eight IRLS iterations. Let $r_n$ denote the perpendicular residual of viewing ray $n$. Its robust scale is estimated as
\begin{equation}
s=1.4826\,\operatorname{median}_n\left|r_n-\operatorname{median}(r)\right|.
\end{equation}
Using cutoff $1.5s$, the weights are updated as
\begin{equation}
w_n=
\begin{cases}
1, & r_n\le1.5s,\\
\dfrac{1.5s}{\max(r_n,10^{-12})}, & \text{otherwise}.
\end{cases}
\end{equation}
If the resulting linear system is poorly conditioned, we fall back to a least-squares solution.

\subsection{Canonical Camera Convention}
The sign of the estimated orbit normal is chosen such that the reconstructed source cameras lie predominantly on its positive side. Radial camera-center vectors are projected onto the orbit plane and sorted by length; the first non-degenerate projected direction defines $\bm e_{i,1}$, and $\bm e_{i,2}=\bm u_i\times\bm e_{i,1}$. The implemented camera-to-world rotation follows a right/down/forward basis convention.

\subsection{Source-Camera Correspondence}
Let $N_i$ be the number of source UAV observations and $K_i$ the number uniquely associated with valid recovered cameras. We require
\begin{equation}
K_i\ge8,\qquad \frac{K_i}{N_i}\ge0.90.
\end{equation}
Matching first uses complete filenames and then extension-stripped stems when necessary. Ambiguous matches are rejected, recovered cameras cannot be assigned to multiple observations, and valid cameras must have finite position and rotation together with positive focal length and image dimensions.

\section{Additional Encoder Details}
\label{sup:encoder}

\subsection{Permutation Invariance}
OCL uses neither observation-index embeddings nor camera-pose embeddings. Let $\bm\Pi$ be a permutation matrix acting on the rows of the concatenated observation-token matrix. Permuting the key and value rows together gives
\begin{equation}
\operatorname{softmax}\left(\bm Q\bm K^\top\bm\Pi^\top\right)(\bm\Pi\bm V)=\operatorname{softmax}\left(\bm Q\bm K^\top\right)\bm V.
\end{equation}
Therefore, up to numerical roundoff, the resulting descriptor is invariant to the ordering of input observations, allowing OCL to process unordered observation sets of variable cardinality.

\section{Evaluation Details}
\label{sup:evaluation}

\subsection{Baseline Provenance}

All trainable comparison methods are retrained on University-1652 using their original or official implementations and are evaluated with the same local query/gallery manifests and retrieval evaluator, while retaining their method-specific preprocessing and descriptor-extraction procedures. The resulting checkpoints are directly evaluated on University-1652, SUES-200, and IRVL328 without further target-domain adaptation. For VFM-Loc$^\dagger$~\cite{lu2026vfmloc}, which is training-free, satellite imagery contributes no fitting statistics; PCA is estimated only from UAV-query descriptors. Ours (NCE-only) uses the same University-1652 source split and optimization budget as OCL but uses only real UAV observations and scene-prototype InfoNCE, without rendered observations or completed-orbit distillation.

\subsection{IRVL328 Query Representation}
The IRVL328 UAV queries in our evaluation copy are stored as 8-bit three-channel images with identical R, G, and B values before preprocessing. The loader applies convert('RGB'), bicubic resizing to 256 pixels, a $256\times256$ center crop, conversion to $[0,1]$ floating-point tensors, and ImageNet normalization. The frozen University-1652-trained encoder is used unchanged, without a thermal-specific encoder, target-domain fine-tuning, or adaptation. The available project artifacts do not specify how the original sensor measurements were converted into the stored grayscale-appearance images, and we therefore make no claim regarding radiometric or temperature calibration.

\subsection{Weather Generation}
\label{sup:weather}

Following~\cite{wang2024multiple,wen2025WeatherPrompt}, weather-style transformations are applied online only to University-1652 UAV queries, while the 951-image satellite gallery remains unchanged and is encoded once. We use imgaug 0.4.0 with fixed seeds for each augmenter. The main transformation settings are summarized in Table~\ref{tab:sup_weather_config}. No augmented images are stored on disk. For reproducibility, the evaluation metadata records the SHA256 hash of the exact transformation implementation used for evaluation.

\begin{table*}[t]
\centering
\small
\setlength{\tabcolsep}{5.0pt}
\renewcommand{\arraystretch}{1.08}
\begin{tabular}{@{}llll@{}}
\toprule
Condition & Operator & Key parameters & Seed(s) \\
\midrule
Fog
& CloudLayer
& mean intensity 225, sparsity 0.9, density 0.5
& 35 \\

Rain
& $5\times$ Rain
& drop size 0.05--0.20, speed 0.04--0.06
& 38, 35, 73, 93, 95 \\

Snow
& $5\times$ Snowflakes
& flake size 0.5--0.9, speed 0.007--0.03
& 38, 35, 74, 94, 96 \\

Fog+Rain
& Fog + Rain
& compound visibility degradation
& 35, 36 \\

Fog+Snow
& Fog + Snow
& compound visibility degradation
& 35, 36 \\

Rain+Snow
& Rain + Snow
& mixed rain and snow effects
& 35, 92, 91, 74 \\

Dark
& Alpha blending + brightness
& $\alpha=0.5$, multiplier 0.2, offset $-30$ to $-15$
& 31, 1991 \\

Over-exposure
& Brightness adjustment
& multiplier 1.6, offset 0--30
& 1992 \\

Wind
& Motion blur
& kernel size 15
& 17 \\
\bottomrule
\end{tabular}

\caption{
Weather-generation settings used for University-1652 robustness evaluation.
All transformations are applied online to UAV queries only, while the satellite gallery remains unchanged.
Rain and snow use fixed multi-stage compositions, and intervals denote the parameter ranges of the corresponding augmenters.
}
\label{tab:sup_weather_config}
\end{table*}

For over-exposure, the additive brightness offset is sampled from $[0,30]$ rather than fixed to a single value. The augmenter seeds are fixed in the implementation, and the exact transformation code used for evaluation is identified by its SHA256 hash in the evaluation metadata.

\subsection{Additional Weather Robustness Analysis}
\label{sup:weather_analysis}

\begin{figure}[t]
    \centering
    \includegraphics[width=\columnwidth]{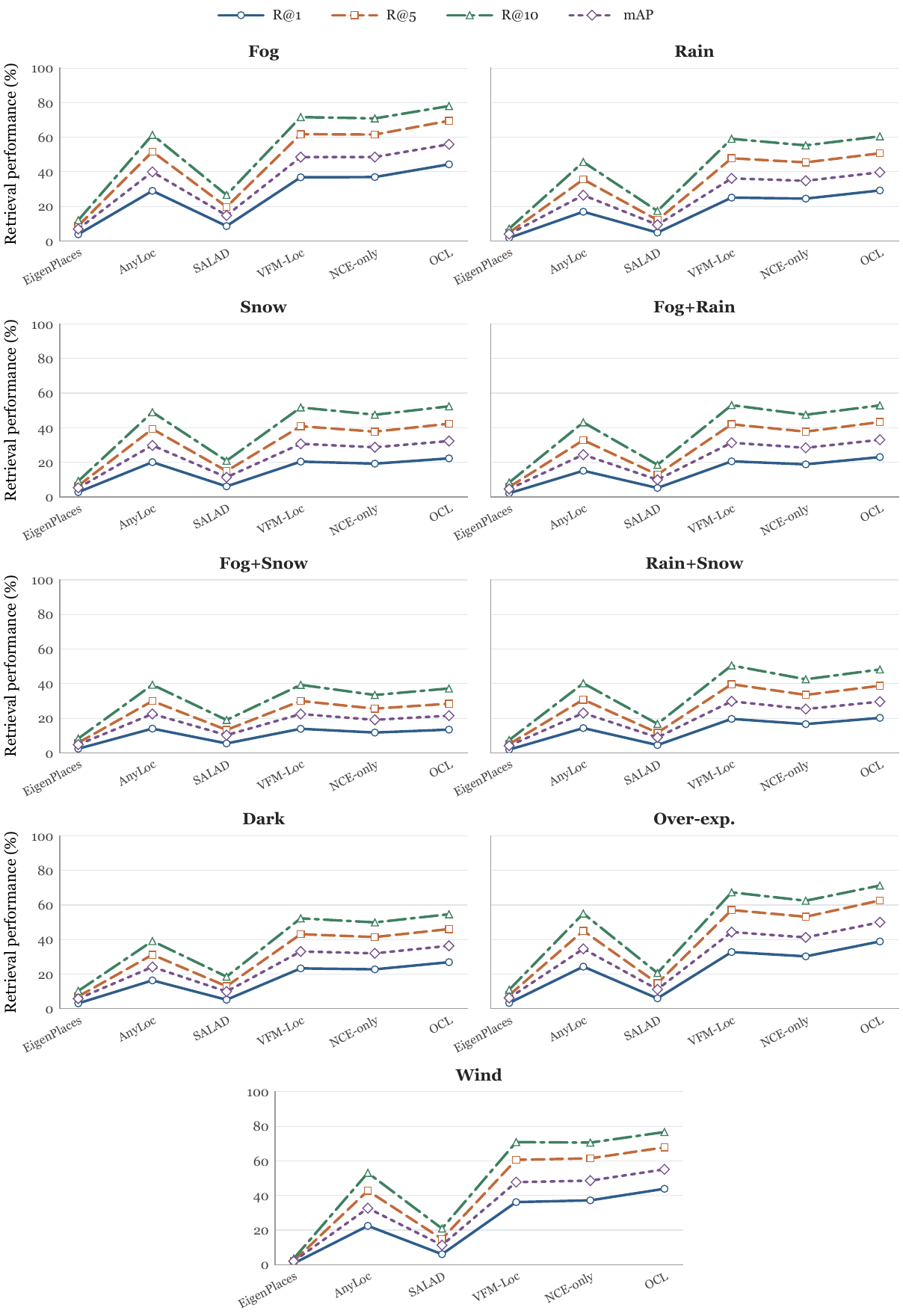}
    \caption{
    Retrieval robustness of target-satellite-free methods under nine weather-style appearance shifts on University-1652.
    Each panel corresponds to one condition and compares R@1, R@5, R@10, and mAP across the methods evaluated in the main paper.
    All panels share the same vertical scale for direct comparison.
    }
    \label{fig:sup_weather_profiles}
\end{figure}

Fig.~\ref{fig:sup_weather_profiles} provides a complementary view of the weather robustness results by comparing all target-satellite-free methods under each appearance shift. Across most conditions, OCL exhibits the strongest overall retrieval profile, with particularly clear advantages under fog, rain, snow, fog+rain, darkness, over-exposure, and wind. The trends across R@1, R@5, R@10, and mAP are largely consistent, showing that the robustness gains extend beyond Top-1 retrieval. The remaining exceptions occur under severe compound shifts, where the advantage becomes smaller, consistent with the quantitative results in the main paper.

\section{Additional Quantitative Results}
\label{sup:quantitative}

\subsection{Extended Retrieval Metrics}
\label{sup:extended_metrics}

The main paper reports R@1 and mAP. For completeness, Table~\ref{tab:sup_extended_metrics} reports the complementary R@5, R@10, and MRR metrics for all evaluated methods under the same protocols and checkpoints. Since each query has a single positive satellite match in our evaluation, MRR is equivalent to mAP; we retain MRR here to explicitly characterize the rank of the correct match.

\begin{table*}[t]
\centering
\small
\renewcommand{\arraystretch}{1.05}

\resizebox{\textwidth}{!}{%
\begin{tabular}{@{}lcccccccccccccccccc@{}}
\toprule
& \multicolumn{3}{c}{University-1652}
& \multicolumn{12}{c}{SUES-200}
& \multicolumn{3}{c}{IRVL328} \\
\cmidrule(lr){2-4}
\cmidrule(lr){5-16}
\cmidrule(lr){17-19}

Method
& \multicolumn{3}{c}{Drone$\rightarrow$Sat.}
& \multicolumn{3}{c}{150\,m}
& \multicolumn{3}{c}{200\,m}
& \multicolumn{3}{c}{250\,m}
& \multicolumn{3}{c}{300\,m}
& \multicolumn{3}{c}{IR$\rightarrow$Vis.} \\
\cmidrule(lr){2-4}
\cmidrule(lr){5-7}
\cmidrule(lr){8-10}
\cmidrule(lr){11-13}
\cmidrule(lr){14-16}
\cmidrule(lr){17-19}

& R@5 & R@10 & MRR
& R@5 & R@10 & MRR
& R@5 & R@10 & MRR
& R@5 & R@10 & MRR
& R@5 & R@10 & MRR
& R@5 & R@10 & MRR \\
\midrule

\multicolumn{19}{@{}l}{\textit{Satellite-accessed before final retrieval}} \\

Instance Loss~\cite{zheng2020dual}
& 77.71 & 84.36 & 66.41
& 55.00 & 67.53 & 38.12
& 57.73 & 70.17 & 43.30
& 57.73 & 69.35 & 44.32
& 58.27 & 68.27 & 45.26
& 15.27 & 21.09 & 10.66 \\

FSRA~\cite{dai2021transformer}
& \best{99.96} & \best{99.97} & \best{99.29}
& 81.12 & 88.60 & 64.80
& 88.17 & 93.83 & 73.33
& 90.97 & 95.38 & 78.30
& 92.83 & 96.12 & 81.61
& 30.33 & 40.57 & 21.38 \\

Sample4Geo~\cite{deuser2023sample4geo}
& 97.55 & 98.13 & 94.57
& \second{89.20} & \second{94.27} & \second{79.12}
& \second{95.88} & \second{98.25} & \second{87.71}
& \second{97.40} & \second{98.60} & \second{91.25}
& \second{98.25} & \second{98.83} & \second{93.60}
& 41.25 & 51.69 & 28.29 \\

Safe-Net~\cite{lin2024self}
& 92.75 & 94.91 & 85.89
& 84.90 & 91.30 & 71.82
& 90.83 & 94.88 & 80.04
& 93.40 & 96.05 & 85.49
& 94.40 & 96.92 & 87.23
& \second{45.18} & \second{57.70} & \second{34.78} \\

MFRGN~\cite{wang2024mfrgn}
& \second{97.65} & \second{98.16} & \second{95.44}
& 87.78 & 92.95 & 75.59
& 93.55 & 96.05 & 84.67
& 95.35 & 97.02 & 89.75
& 95.80 & 97.92 & 91.15
& 37.03 & 46.93 & 28.52 \\

WeatherPrompt~\cite{wen2025WeatherPrompt}
& 93.16 & 95.55 & 87.27
& 78.62 & 86.83 & 64.91
& 88.50 & 94.27 & 75.86
& 93.77 & 97.17 & 82.73
& 95.95 & 98.10 & 87.47
& 37.49 & 50.11 & 26.22 \\

SpikeViMFormer~\cite{chen2026efficient}
& 96.73 & 97.89 & 91.89
& 88.83 & 94.12 & 78.91
& 93.33 & 96.92 & 86.22
& 95.10 & 97.22 & 89.26
& 95.88 & 97.80 & 90.99
& 40.92 & 49.59 & 29.02 \\

EM-CVGL~\cite{li2024learning}
& 91.03 & 94.98 & 79.42
& 86.33 & 92.33 & 72.46
& 93.60 & 96.05 & 81.87
& 95.93 & 97.10 & 88.12
& 96.73 & 97.95 & 91.44
& 32.53 & 45.53 & 21.49 \\

UCVGL~\cite{li2024unleashing}
& 53.37 & 61.09 & 44.89
& 40.40 & 53.00 & 31.38
& 51.15 & 62.20 & 40.56
& 55.55 & 66.43 & 45.50
& 60.55 & 69.35 & 49.41
& 21.28 & 28.58 & 15.82 \\

DMNIL~\cite{chen2026without}
& 96.38 & 97.38 & 92.61
& \best{94.20} & \best{96.73} & \best{84.41}
& \best{97.72} & \best{98.47} & \best{91.41}
& \best{98.70} & \best{99.12} & \best{94.54}
& \best{99.30} & \best{99.62} & \best{96.47}
& \best{67.23} & \best{77.78} & \best{49.78} \\

\midrule

\multicolumn{19}{@{}l}{\textit{Target-satellite-free}} \\

EigenPlaces~\cite{berton2023eigenplaces}
& 9.94 & 13.38 & 7.86
& 26.55 & 35.73 & 20.03
& 32.25 & 40.42 & 24.01
& 34.77 & 42.85 & 27.17
& 35.98 & 43.73 & 29.01
& 8.52 & 12.84 & 6.90 \\

AnyLoc~\cite{keetha2023anyloc}
& 33.02 & 42.24 & 25.49
& 62.73 & 74.10 & 48.09
& 67.35 & 76.90 & 53.46
& 72.98 & 81.30 & 58.64
& 75.20 & 84.05 & 60.77
& 19.77 & 29.16 & 13.20 \\

SALAD~\cite{izquierdo2024optimal}
& 18.84 & 25.58 & 14.34
& 30.95 & 41.75 & 24.12
& 37.10 & 47.48 & 30.30
& 44.07 & 54.00 & 34.93
& 49.85 & 59.33 & 40.13
& 11.86 & 18.48 & 9.09 \\

VFM-Loc$^\dagger$~\cite{lu2026vfmloc}
& 57.79 & 67.96 & 44.90
& \second{64.20} & 75.38 & 48.47
& \second{75.40} & \second{86.38} & 58.33
& \best{83.95} & \best{91.05} & \second{67.48}
& \best{89.88} & \best{93.60} & \best{77.70}
& 19.28 & 29.52 & 12.75 \\

Ours (NCE-only)
& \second{74.59} & \second{82.10} & \second{61.05}
& 62.48 & \second{75.68} & \second{49.38}
& 73.18 & 82.90 & \second{59.83}
& 78.10 & 87.15 & 66.06
& 82.28 & 89.25 & 72.74
& \second{19.91} & \second{30.14} & \second{13.37} \\

\textbf{Ours (OCL)}
& \best{80.98} & \best{87.42} & \best{68.44}
& \best{68.33} & \best{81.00} & \best{57.23}
& \best{76.53} & \best{88.20} & \best{65.01}
& \second{81.18} & \second{90.03} & \best{69.81}
& \second{83.65} & \second{90.45} & \second{74.60}
& \best{22.81} & \best{33.39} & \best{15.11} \\

\bottomrule
\end{tabular}%
}

\caption{
Extended cross-benchmark drone-to-satellite retrieval results (\%).
Only the additional metrics omitted from the main paper are reported.
\textcolor{red}{\textbf{Red}} and \textcolor{blue}{\textbf{blue}} denote the best and second-best results within each protocol group for each metric, respectively.
Since each query has a single positive satellite match, MRR is equivalent to mAP.
$^\dagger$VFM-Loc uses UAV-query-only PCA in our evaluation.
}
\label{tab:sup_extended_metrics}
\end{table*}

The extended metrics are consistent with the main results. On University-1652, OCL ranks first among all target-satellite-free methods in R@5, R@10, and MRR. The same pattern holds on IRVL328, where OCL improves over NCE-only from 19.91/30.14 to 22.81/33.39 in R@5/R@10, indicating better ranking quality under the infrared-to-visible modality shift.

SUES-200 exhibits a consistent altitude-dependent trend. OCL outperforms the architecture-matched NCE-only baseline across all four altitudes in R@5, R@10, and MRR. At 150\,m, OCL improves R@5/R@10/MRR from 62.48/75.68/49.38 to 68.33/81.00/57.23. The gains remain consistent at 200\,m (76.53/88.20/65.01 vs.\ 73.18/82.90/59.83) and 250\,m (81.18/90.03/69.81 vs.\ 78.10/87.15/66.06). At 300\,m, OCL further reaches 83.65\% R@5, 90.45\% R@10, and 74.60\% MRR, compared with 82.28/89.25/72.74 for NCE-only. The improvement is largest at lower altitudes and gradually narrows as the UAV viewpoint becomes more global, consistent with the main-paper observation that orbit completion is particularly helpful under larger drone--satellite viewpoint discrepancies.

\subsection{Extended Controlled Ablation Metrics}
\label{sup:extended_ablation}

\begin{table}[t]
\centering
\small
\setlength{\tabcolsep}{4.0pt}
\renewcommand{\arraystretch}{1.06}

\begin{tabular}{@{}lcccc@{}}
\toprule
Setting & R@1 & R@5 & R@10 & mAP \\
\midrule

\multicolumn{5}{@{}l}{\textit{(a) Student and teacher completion}} \\

Real only
& 51.17 & 76.07 & 83.66 & 62.55 \\

Teacher completion
& 53.24 & 77.52 & 84.87 & 64.39 \\

Student augmentation
& 55.71 & 79.39 & 86.00 & 66.57 \\

\textbf{Full OCL}
& \textbf{57.85} & \textbf{80.98} & \textbf{87.42} & \textbf{68.44} \\

\midrule

\multicolumn{5}{@{}l}{\textit{(b) Geometry versus appearance}} \\

Real context
& 51.17 & 76.07 & 83.66 & 62.55 \\

+ Appearance
& 50.51 & 75.56 & 83.10 & 61.96 \\

+ Canonical
& \textbf{53.24} & \textbf{77.52} & \textbf{84.87} & \textbf{64.39} \\

\midrule

\multicolumn{5}{@{}l}{\textit{(c) Training objectives}} \\

\textbf{Full OCL}
& \textbf{57.85} & \textbf{80.98} & \textbf{87.42} & \textbf{68.44} \\

w/o Distillation
& 55.18 & 78.63 & 85.34 & 65.88 \\

w/o InfoNCE
& 36.45 & 61.16 & 71.80 & 48.14 \\

\bottomrule
\end{tabular}

\caption{
Extended retrieval metrics for the controlled ablations on University-1652 (\%).
The configurations, checkpoints, and evaluation protocol are identical to those used in Table~\ref{tab:ablation_main}; R@5 and R@10 are additionally reported here to characterize retrieval quality beyond rank 1.
}
\label{tab:sup_extended_ablation}
\end{table}

To complement the R@1 and mAP results reported in the main paper, we report R@5 and R@10 for the same controlled ablation runs in Table~\ref{tab:sup_extended_ablation}. All results are obtained from the same configurations, checkpoints, and evaluation protocol as the corresponding experiments in Table~\ref{tab:ablation_main}. This extended evaluation therefore directly characterizes whether the effects observed at rank 1 remain consistent over a broader retrieval range.\\
\textbf{Student augmentation and teacher completion.}
The higher-rank retrieval metrics follow the same trend as R@1 and mAP. Starting from the real-only configuration, adding canonical views only to the teacher increases R@5/R@10 from 76.07/83.66 to 77.52/84.87, while student-side augmentation alone reaches 79.39/86.00. Combining both components further improves performance to 80.98\% R@5 and 87.42\% R@10. Relative to student-side augmentation alone, completed-orbit teacher supervision therefore provides additional gains of 1.59 and 1.42 points in R@5 and R@10, respectively. These results confirm that the teacher contributes beyond using rendered views merely as additional student training samples.\\
\textbf{Geometry versus appearance augmentation.}
When the student is restricted to real UAV observations, conventional appearance augmentation slightly reduces R@5/R@10 from 76.07/83.66 to 75.56/83.10. In contrast, replacing the appearance-augmented teacher context with canonical 3DGS views increases R@5/R@10 to 77.52/84.87. Compared with appearance augmentation, canonical views improve R@5 and R@10 by 1.96 and 1.77 points, respectively. Together with the corresponding R@1 and mAP improvements, these results indicate that the benefit of teacher completion arises from complementary geometric viewpoint coverage rather than simply increasing the amount or appearance diversity of teacher observations.\\
\textbf{Training objectives.}
The complementary roles of distillation and InfoNCE are also evident at higher retrieval ranks. Removing distillation decreases R@5/R@10 from 80.98/87.42 to 78.63/85.34, corresponding to drops of 2.35 and 2.08 points, respectively. Removing InfoNCE causes a much larger degradation to 61.16\% R@5 and 71.80\% R@10, with drops of 19.82 and 15.62 points. This behavior is consistent with the R@1 and mAP results in the main paper: scene-level contrastive discrimination provides the primary signal for place retrieval, while partial-to-completed distillation transfers additional completed-orbit context to improve ranking quality.

\section{Partial-Orbit Robustness}
\label{sup:partial_orbit}

Standard drone-to-satellite retrieval uses a single UAV image per query, but OCL naturally supports a variable number of observations. We evaluate a fixed OCL checkpoint with $m\in\{1,2,4,8,16\}$ UAV observations per test location on University-1652 using place-balanced nested query sets, such that each larger set contains all observations from the preceding one.

\begin{figure}[t]
    \centering
    \includegraphics[width=\columnwidth]{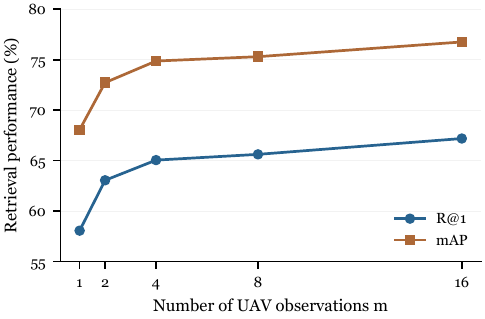}
    \caption{
    Retrieval performance with increasing numbers of UAV observations on University-1652.
    Both R@1 and mAP improve consistently as additional viewpoints become available, with the largest gain occurring from one to two observations and diminishing returns thereafter.
    }
    \label{fig:sup_partial_orbit}
\end{figure}

As shown in Table~\ref{tab:sup_partial_orbit} and Fig.~\ref{fig:sup_partial_orbit}, retrieval performance improves monotonically as additional UAV observations become available. The largest gain occurs from one to two views, where R@1 increases from 58.06\% to 63.05\% and mAP from 68.06\% to 72.73\%. Further observations yield consistent but diminishing gains, reaching 67.19\% R@1 and 76.76\% mAP with 16 views. This trend indicates that a small amount of additional viewpoint coverage already provides substantial complementary scene information, while later views contribute increasingly redundant context. Importantly, OCL retains the standard single-image operating point while naturally exploiting additional UAV observations when available.

\begin{table}[t]
\centering
\small
\setlength{\tabcolsep}{4.2pt}
\begin{tabular}{lccccc}
\toprule
Views $m$ & 1 & 2 & 4 & 8 & 16\\
\midrule
R@1 & 58.06 & 63.05 & 65.05 & 65.62 & 67.19\\
mAP & 68.06 & 72.73 & 74.87 & 75.30 & 76.76\\
\bottomrule
\end{tabular}
\caption{
Place-balanced drone-to-satellite retrieval (\%) with increasing numbers of UAV observations on University-1652.
Each query set is nested within the next larger one.
The $m{=}1$ setting uses one fixed query per location and is therefore not directly comparable with the image-level main benchmark.
}
\label{tab:sup_partial_orbit}
\end{table}

\section{Training-Time Geometry: Reliability and Cost}
\label{sup:geometry_cost}

OCL uses SfM and 3DGS only during source-side training to construct canonical observations. All University-1652 source scenes successfully complete the reconstruction and 16-view rendering pipeline. As shown in Table~\ref{tab:sup_geometry_diag}, held-out novel-view rendering achieves 26.39\,dB PSNR, 0.8669 SSIM, and 0.0830 LPIPS, indicating sufficient geometric and appearance consistency for generating complementary training viewpoints. The complete geometry preprocessing pipeline, including SfM reconstruction, 3DGS optimization, and canonical-view rendering, requires 27.56 hours and is performed only once before representation learning. The subsequent OCL representation-learning stage takes approximately 3\,h\,48\,min on a single RTX 4090 with 3.21\,GB peak PyTorch-allocated GPU memory. After training, SfM, 3DGS, canonical renders, the EMA teacher, and prototype memory are discarded, so the geometric pipeline introduces no inference-time overhead.

\begin{figure}[t]
    \centering
    \includegraphics[width=\columnwidth]{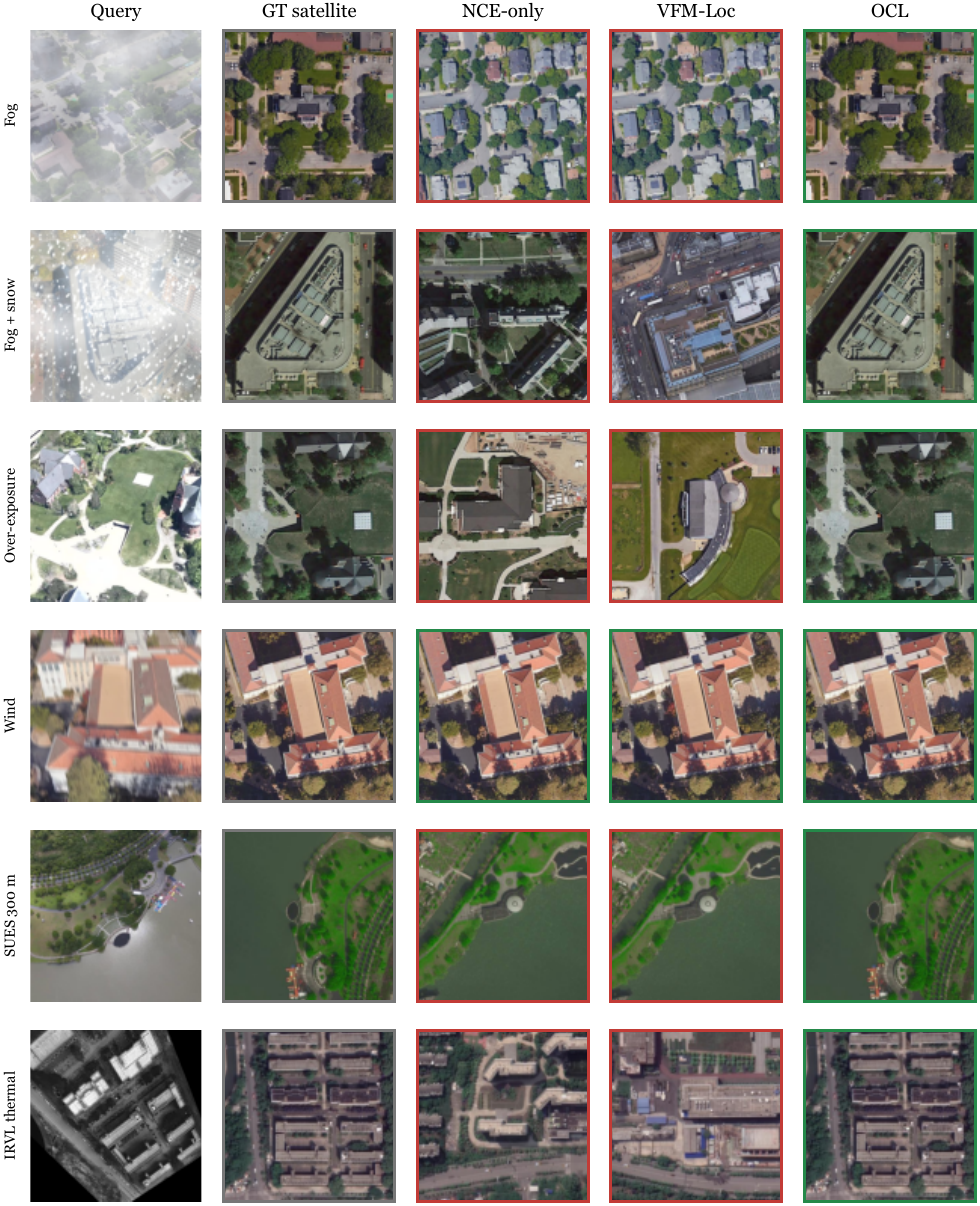}
    \caption{
    Additional qualitative retrieval examples under unseen appearance, viewpoint, and modality shifts.
    Columns show the UAV query, ground-truth satellite image, and Top-1 retrievals from Ours (NCE-only), VFM-Loc, and OCL.
    Green and red borders denote correct and incorrect retrievals, respectively.
    The examples include weather-corrupted University-1652 queries, SUES-200 at 300\,m, and IRVL328 infrared queries.
    }
    \label{fig:sup_generalization_retrieval}
\end{figure}

\begin{table}[t]
\centering
\small
\setlength{\tabcolsep}{5.0pt}
\begin{tabular}{lc}
\toprule
Diagnostic & Value\\
\midrule
Held-out PSNR & 26.39 dB\\
Held-out SSIM & 0.8669\\
Held-out LPIPS & 0.0830\\
\midrule
Geometry preprocessing & 27.56 h\\
Representation learning & 3 h 48 min\\
Peak training memory & 3.21 GB\\
\bottomrule
\end{tabular}
\caption{Rendering quality and training-time computational cost on University-1652. Geometry preprocessing includes SfM, 3DGS optimization, and canonical-view rendering for all source scenes.}
\label{tab:sup_geometry_diag}
\end{table}

\section{Additional Qualitative Results}
\label{sup:qualitative}

\subsection{Additional Retrieval Examples}
\label{sup:retrieval_examples}

Fig.~\ref{fig:sup_generalization_retrieval} provides additional qualitative retrieval examples under appearance, viewpoint, and modality shifts. We compare OCL with Ours (NCE-only) and VFM-Loc on weather-corrupted University-1652 queries, SUES-200 at 300\,m, and IRVL328. Across fog, fog+snow, over-exposure, SUES-200, and IRVL328, the baselines frequently retrieve visually plausible but incorrect satellite locations, whereas OCL recovers the ground-truth location in the illustrated examples. This behavior is particularly evident when local appearance cues are strongly corrupted or when the query undergoes a substantial viewpoint or modality shift, suggesting that completed-orbit supervision encourages greater reliance on scene-level spatial structure. The wind example further shows a case in which all three methods succeed, indicating that the qualitative comparison is not restricted to cases where only OCL is correct.

\subsection{Failure Cases}
\label{sup:failure_cases}

\begin{figure}[t]
    \centering
    \includegraphics[width=\columnwidth]{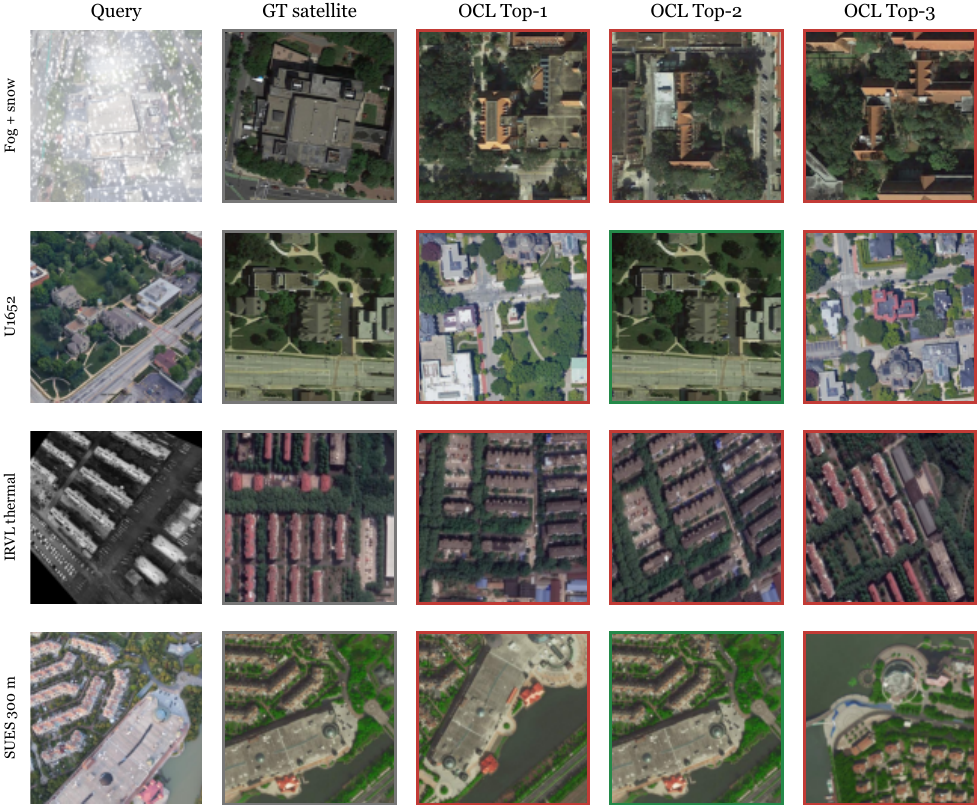}
    \caption{
    Representative OCL failure cases.
    Each row shows the UAV query, ground-truth satellite image, and the Top-3 OCL retrievals.
    Red borders denote incorrect retrievals, while a green border indicates the ground-truth location when it appears within the Top-3.
    The examples cover compound weather corruption, University-1652 appearance ambiguity, IRVL328 infrared-to-visible transfer, and SUES-200 viewpoint shift.
    }
    \label{fig:sup_failure_cases}
\end{figure}

Figure~\ref{fig:sup_failure_cases} presents representative cases where OCL fails to rank the correct satellite image first. The examples span severe compound appearance corruption, ambiguous University-1652 scenes, infrared-to-visible retrieval, and cross-dataset viewpoint changes on SUES-200. In several cases, the highest-ranked predictions share coarse layout or land-cover patterns with the ground truth but differ in local spatial organization, illustrating that scene-level structural similarity can still be ambiguous when discriminative details are weak or corrupted. Under fog+snow and infrared queries, substantial appearance degradation further reduces the correspondence between UAV and satellite observations. The SUES-200 example similarly illustrates the remaining difficulty of large viewpoint changes. These cases indicate that orbit completion improves robustness but cannot fully resolve retrieval when the query provides insufficient or ambiguous evidence.

\subsection{Held-Out Rendering Examples}
\label{sup:rendering_examples}

\begin{figure}[t]
    \centering
    \includegraphics[width=\columnwidth]{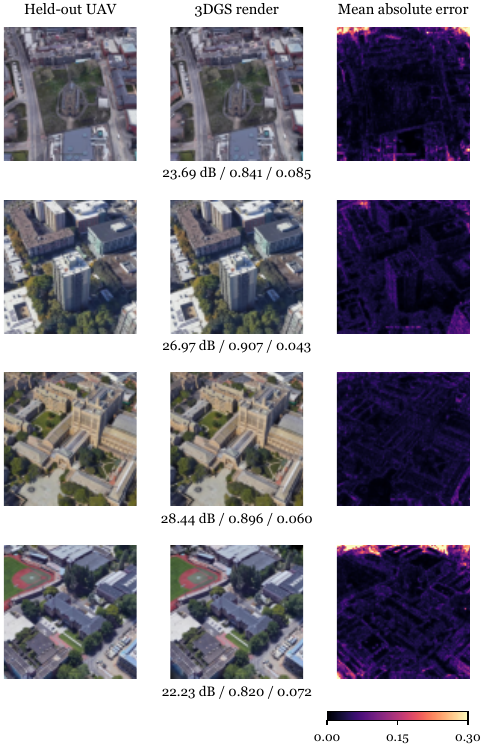}
    \caption{
    Held-out novel-view rendering examples from the source-side geometry pipeline.
    Each row shows a real UAV target, its corresponding 3DGS rendering, and the mean absolute RGB error.
    The values below the rendered images report PSNR/SSIM/LPIPS.
    All error maps use the same range of $[0,0.30]$ for direct comparison.
    These reconstructions are used only to generate privileged training observations and are discarded at inference.
    }
    \label{fig:sup_rendering_quality}
\end{figure}

Fig.~\ref{fig:sup_rendering_quality} provides qualitative examples from the held-out novel-view evaluation used to assess the source-side geometry pipeline. Each row compares a real UAV image excluded from 3DGS appearance optimization with the corresponding rendering from the recovered geometry and its mean absolute RGB error. The examples span different reconstruction qualities, with PSNR ranging from 22.23 to 28.44\,dB.

Overall, the rendered views preserve the dominant building layout, road structure, and scene geometry across the examples. The error maps show that larger discrepancies are concentrated around fine structures, high-frequency boundaries, and local appearance details, while the overall spatial organization remains well aligned. Importantly, OCL does not require photorealistic novel-view synthesis: these renderings serve only as complementary training observations for orbit completion. The geometric pipeline is discarded after training, so rendering artifacts do not directly propagate to inference-time satellite retrieval.

\end{document}


\maketitle

\section{Overview}
This supplement expands the method, implementation, and experiments reported in the main paper. All ablation results use the fixed seed 0; we do not make an across-seed variance or statistical-significance claim. The final University-1652 result in the main paper (57.85\% R@1 / 68.44\% mAP) is produced by the selected \texttt{train\_u1652} student checkpoint. The M8 result in the ablation table (57.84/68.44) is a separate run from the progressive ablation branch. SUES-200 and IRVL328 results are \emph{not} cross-dataset zero-shot transfer: OCL is trained separately from each benchmark's UAV training split (and separately for each SUES altitude) while satellite imagery remains unavailable during task-specific development.

\section{Detailed Method}
\subsection{Source-scene reconstruction}
For source scene $i$, observed UAV images are denoted
\begin{equation}
\R_i=\{I^r_{i,n}\}_{n=1}^{n_i}.
\end{equation}
COLMAP first recovers camera intrinsics, extrinsics, and sparse structure from overlapping source observations~\cite{schoenberger2016sfm}. Let successfully registered camera centers and camera-to-world rotations be $\bm c_{i,n}\in\mathbb R^3$ and $\bm Q_{i,n}\in SO(3)$. These assets initialize a 3D Gaussian Splatting (3DGS) representation~\cite{kerbl3Dgaussians}
\begin{equation}
\mathcal G_i=\{(\bm\mu_g,\bm\Sigma_g,\alpha_g,\bm a_g)\}_{g=1}^{G_i},
\end{equation}
where $\bm\mu_g$ and $\bm\Sigma_g$ are Gaussian position and covariance, $\alpha_g$ is opacity, and $\bm a_g$ contains spherical-harmonic appearance coefficients. The reconstruction objective is the standard mixture of $L_1$ and DSSIM terms,
\begin{equation}
\mathcal L_{\mathrm{GS}}=0.8\|\hat I-I\|_1+0.2\bigl(1-\mathrm{SSIM}(\hat I,I)\bigr).
\end{equation}
The downstream OCL loss does not back-propagate into the reconstruction pipeline.

The orbit-completion assets used by OCL are rendered from the iteration-7000 Gaussian snapshot. For the dedicated 30-scene University-1652 reconstruction benchmark in Sec.~\ref{sec:geometry_diag}, 3DGS is explicitly optimized for 7,000 iterations. Exposure compensation is not enabled in the evaluated configuration. The historical SUES assets contain longer training logs in some cases, but OCL uses their saved iteration-7000 snapshot; therefore we report the snapshot used by OCL rather than treating a longer historical log as part of the representation-learning method.

\subsection{Robust scene target and canonical orbit}
Under our camera convention, the third column of $\bm Q_{i,n}$ is the camera optical axis,
\begin{equation}
\bm d_{i,n}=\bm Q_{i,n}\bm e_3,\qquad \|\bm d_{i,n}\|_2=1.
\end{equation}
We estimate the source-scene target as the robust point closest to all reconstructed viewing \emph{lines},
\begin{equation}
\bm t_i=\arg\min_{\bm t}\sum_{n\in\mathcal J_i}w_{i,n}
\left\|(\bm I-\bm d_{i,n}\bm d_{i,n}^{\top})(\bm t-\bm c_{i,n})\right\|_2^2.
\label{eq:supp_target}
\end{equation}
The weights are updated by IRLS for at most eight iterations. If $r_n$ denotes the current perpendicular line residual, the robust scale is $1.4826\,\mathrm{median}(|r_n-\mathrm{median}(r)|)$ and the cutoff is $1.5$ times this scale. Residuals below the cutoff receive unit weight; larger residuals receive cutoff$/\max(r_n,10^{-12})$. This target calculation imposes no metric-scale or external georeferencing requirement.

With $\bm r_{i,n}=\bm c_{i,n}-\bm t_i$, the centered camera-position covariance is
\begin{equation}
\bm S_i=\frac{1}{|\mathcal J_i|}\sum_{n\in\mathcal J_i}
(\bm r_{i,n}-\bar{\bm r}_i)(\bm r_{i,n}-\bar{\bm r}_i)^\top.
\end{equation}
The canonical up axis $\bm u_i$ is the unit eigenvector of $\bm S_i$ with the smallest eigenvalue. Its sign is selected so that the median camera displacement along the up axis is non-negative, with an optical-axis tie breaker for near-degenerate cases. Camera radial vectors are projected into the orbit plane; among them, the longest non-degenerate projected direction deterministically defines $\bm e_{i,1}$, and
\begin{equation}
\bm e_{i,2}=\bm u_i\times\bm e_{i,1}
\end{equation}
completes a right-handed scene-specific basis. The orbit radius is
\begin{equation}
\rho_i=\operatorname{median}_{n\in\mathcal J_i}\|\bm c_{i,n}-\bm t_i\|_2.
\end{equation}

For azimuth $\theta$ and tilt $\phi$ ($\phi=0^\circ$ is nadir),
\begin{align}
\bm h_i(\theta)&=\sin\theta\,\bm e_{i,1}+\cos\theta\,\bm e_{i,2},\\
\bm f_i(\theta,\phi)&=\sin\phi\,\bm h_i(\theta)-\cos\phi\,\bm u_i,\\
\bm c_i^c(\theta,\phi)&=\bm t_i-\rho_i\bm f_i(\theta,\phi),\\
\bm r_i(\theta)&=\cos\theta\,\bm e_{i,1}-\sin\theta\,\bm e_{i,2}.
\end{align}
The camera down direction is $\bm f_i\times\bm r_i$. We render the Cartesian product
\begin{equation}
\mathcal T=\{35^\circ,25^\circ,15^\circ,5^\circ\},\qquad
\mathcal H=\{0^\circ,90^\circ,180^\circ,270^\circ\},
\end{equation}
for $K_c=16$ complementary views. Render width, height, horizontal FOV, and vertical FOV are inherited from the per-scene medians of the reconstructed source cameras. The completed orbit is
\begin{equation}
\Oset_i=\R_i\cup\C_i.
\end{equation}

\subsection{ViewTokenizer}
A DINOv3 ViT-B/16 backbone~\cite{simeoni2025dinov3} processes a $256\times256$ image into $P=256$ patch tokens $\bm X\in\mathbb R^{P\times768}$ after removing prefix tokens. Four learned queries $\bm V^{(0)}\in\mathbb R^{4\times768}$ compress each image. For $\ell\in\{1,2\}$,
\begin{align}
\widetilde{\bm V}^{(\ell)}&=\bm V^{(\ell-1)}+
\mathrm{MHA}\bigl(\mathrm{LN}_{q,\ell}(\bm V^{(\ell-1)}),
\mathrm{LN}_{x,\ell}(\bm X),\mathrm{LN}_{x,\ell}(\bm X)\bigr),\\
\bm V^{(\ell)}&=\widetilde{\bm V}^{(\ell)}+
\mathrm{FFN}_{\ell}(\mathrm{LN}_{f,\ell}(\widetilde{\bm V}^{(\ell)})).
\end{align}
Both blocks use 8 attention heads and a 3072-dimensional GELU feed-forward layer. The resulting image tokens are
\begin{equation}
T_\theta(I)=\bm V^{(2)}\in\mathbb R^{4\times768}.
\end{equation}
The four queries have no predefined semantic roles.

\subsection{OrbitFormer and order invariance}
For $m$ observations, their image tokens are concatenated,
\begin{equation}
\bm U=\mathrm{Concat}_{j=1}^{m}T_\theta(I_j)\in\mathbb R^{4m\times768}.
\end{equation}
OrbitFormer maintains a PLACE query and eight additional latents,
\begin{equation}
\bm Z^{(0)}=[\bm z_{\mathrm{place}};\bm z_1;\ldots;\bm z_8]\in\mathbb R^{9\times768}.
\end{equation}
A latent-to-view cross-attention block reads $\bm U$,
\begin{equation}
\bm Z^{(1)}=\bm Z^{(0)}+
\mathrm{MHA}(\mathrm{LN}_q(\bm Z^{(0)}),\mathrm{LN}_v(\bm U),\mathrm{LN}_v(\bm U)),
\end{equation}
followed by three latent self-attention/FFN blocks. The normalized PLACE output is the descriptor $z=f_\theta(\{I_j\}_{j=1}^m)$.

No observation-index or camera-pose embedding is added. If the rows of the observation token set are permuted by $\bm\Pi$, the key and value matrices are permuted together and
\begin{equation}
\mathrm{softmax}(\bm Q\bm K^\top\bm\Pi^\top)(\bm\Pi\bm V)
=\mathrm{softmax}(\bm Q\bm K^\top)\bm V.
\end{equation}
Thus OrbitFormer is invariant to the \emph{ordering} of the observations, not to their visual content or viewpoint.

\section{Training Implementation}
\subsection{EMA teacher and completed-orbit prototype}
The teacher is initialized as a copy of the student and receives no gradient. After each successful student optimizer update,
\begin{equation}
\bar\theta_{t+1}=\alpha_t\bar\theta_t+(1-\alpha_t)\theta_{t+1},
\end{equation}
where $\alpha_t$ follows a cosine schedule from 0.996 to 0.9999. For the final mean-teacher formulation, each completed-orbit observation is independently encoded through the teacher singleton path,
\begin{equation}
\bar{\bm z}_{i,v}=f_{\bar\theta}(\{I_{i,v}\}),\qquad
\|\bar{\bm z}_{i,v}\|_2=1,
\end{equation}
and the conceptual completed-orbit target is
\begin{equation}
\bm c_i=\sg\left[\Norm\left(\frac{1}{|\Oset_i|}
\sum_{I_{i,v}\in\Oset_i}\bar{\bm z}_{i,v}\right)\right].
\label{eq:supp_proto}
\end{equation}
All observations have equal weight, including observed and rendered views.

\subsection{Token memory and epoch prototype bank}
Directly re-encoding every completed orbit from pixels at every optimization step is unnecessary. For scene $i$, OCL stores cached EMA-teacher ViewTokenizer outputs
\begin{equation}
\Mmem_i\in\mathbb R^{|\Oset_i|\times4\times768}.
\end{equation}
The memory is initialized by a complete teacher encoding before the epoch loop. It is then maintained at two time scales:
\begin{enumerate}[leftmargin=*]
    \item \textbf{Epoch-start hard refresh.} About 20\% of all observation entries form one rotating shard and are overwritten by freshly encoded teacher tokens. Five near-equal shards are cycled across epochs.
    \item \textbf{Online momentum update.} Before each student optimization step, the entries corresponding to the sampled observations are updated as
    \begin{equation}
    \Mmem_{i,v}\leftarrow0.9\Mmem_{i,v}+0.1\bar{\bm V}_{i,v}.
    \end{equation}
\end{enumerate}
After the epoch-start hard refresh, the current EMA-teacher OrbitFormer independently maps all cached observations through its singleton path and rebuilds one detached prototype for every source-training scene. This $N\times768$ prototype bank is held fixed for the entire epoch. Therefore, the positive prototype and all negatives used within an epoch have a common prototype-construction time, although the cached observation tokens underlying different prototypes may have different ages. The bank is rebuilt again at the next epoch boundary.

The formal training order for the final mean-teacher formulation is summarized below.
\begin{verbatim}
initialize student
teacher <- copy(student)
initialize token memory with teacher ViewTokenizer
for epoch:
    hard-refresh rotating 20% token shard
    train_bank <- rebuild all training scene prototypes
    val_bank   <- rebuild all validation scene prototypes
    for step:
        sample scenes and partial-orbit observations
        momentum-update token memory for sampled observations
        z <- student(partial observations)
        loss <- cosine_distill(z, train_bank[pos])
                + InfoNCE(z, train_bank)
        optimizer step
        if optimizer step succeeds:
            EMA-update teacher
            scheduler step
    source-only validation with fixed val_bank
\end{verbatim}
This implementation is a slowly refreshed approximation to Eq.~\eqref{eq:supp_proto}; neither token memory nor the prototype bank is used after training.

\subsection{Objective and all-scene negatives}
For a logical batch of $B$ source scenes,
\begin{equation}
\mathcal L_{\mathrm{distill}}=\frac{1}{B}\sum_i(1-\bm z_i^\top\bm c_i).
\end{equation}
The detached prototype bank contains every source-training scene,
\begin{equation}
\mathcal P=\{\bm c_j\}_{j\in\mathcal I_{\mathrm{tr}}},
\end{equation}
and InfoNCE is
\begin{equation}
\mathcal L_{\mathrm{NCE}}=-\frac{1}{B}\sum_i
\log\frac{\exp(\bm z_i^\top\bm c_i/\tau)}
{\sum_{j\in\mathcal I_{\mathrm{tr}}}\exp(\bm z_i^\top\bm c_j/\tau)},
\qquad\tau=0.07.
\end{equation}
For the University-1652 development split, the denominator contains 631 prototypes (one positive and 630 negatives). The final loss is
\begin{equation}
\mathcal L=\mathcal L_{\mathrm{distill}}+\mathcal L_{\mathrm{NCE}}.
\end{equation}

\subsection{Observation curriculum}
The student is progressively exposed to fewer observations. The nominal observation-count curriculum is:
\begin{table}[h]
\centering
\small
\begin{tabular}{lccccc}
\toprule
Training progress & $m=16$ & $m=8$ & $m=4$ & $m=2$ & $m=1$\\
\midrule
0--25\% & .40 & .35 & .15 & .05 & .05\\
25--60\% & -- & .25 & .30 & .25 & .20\\
60--100\% & -- & -- & .15 & .30 & .55\\
\bottomrule
\end{tabular}
\caption{Student observation-count curriculum. During the final 40\% of training, a batch-level constraint guarantees that at least half of the scene samples are singletons; with a logical batch of eight places this yields four or five singleton samples.}
\label{tab:supp_curriculum}
\end{table}
For the final mixed-student configuration, sampling chooses the observed-only pool with probability 0.7 and the observed-plus-rendered pool with probability 0.3. Sampling within the chosen pool is without replacement. If $m>1$ and a mixed-pool sample contains only rendered observations, one observation is replaced by an observed source image. For $m=1$, a rendered singleton is permitted. The controlled M1--M3 branch overrides this policy with observed-only student inputs.

\subsection{Optimization and model selection}
\begin{table}[h]
\centering
\small
\begin{tabular}{@{}ll@{}}
\toprule
Setting & Value\\
\midrule
Backbone & DINOv3 ViT-B/16\\
Input resolution & $256\times256$\\
Descriptor dimension & 768\\
ViewTokenizer queries & 4\\
OrbitFormer latents & 9 (1 PLACE + 8 auxiliary)\\
Training length & 20 epochs $\times$ 100 steps\\
Optimizer & AdamW\\
Backbone learning rate & $10^{-6}$\\
New-module learning rate & $10^{-4}$\\
Weight decay & 0.05\\
Layer-wise LR decay & 0.80\\
LR schedule & 5\% warmup + cosine decay\\
Logical batch / microbatch & 8 places / 2 places\\
Gradient clipping & 1.0\\
Contrastive temperature & 0.07\\
Distillation / NCE weights & 1 / 1\\
EMA coefficient & cosine, 0.996 $\rightarrow$ 0.9999\\
Token-memory momentum & 0.9\\
Epoch hard-refresh fraction & 0.20\\
\bottomrule
\end{tabular}
\caption{Representation-learning hyperparameters.}
\label{tab:supp_hparams}
\end{table}

University-1652 source scene IDs are natural-sorted, permuted by NumPy \texttt{default\_rng(0)}, and divided into 70 source-validation scenes and 631 optimization scenes. Checkpoint selection never uses target satellite retrieval. Validation ranks single-image source queries against detached source-scene teacher prototypes. The primary selection statistic is source-validation R@1, followed by MRR, hard-negative margin, and the earlier epoch as successive tie breakers. In configurations with both observed and closure validation tasks, the task-level metrics are averaged with equal weight rather than pooled by query count.

M1--M3 use observed-only validation queries. In the progressive branch, M4 uses observed-only validation; M5--M8 use the saved observed-plus-closure validation configuration. The M4$\rightarrow$M5 transition therefore changes several components and is interpreted as a progressive-system difference rather than a strict one-variable ablation.

\subsection{Deployment boundary}
The exported and evaluated model is the selected \emph{student}, not the EMA teacher. UAV queries and satellite gallery images share exactly the same singleton path
\begin{equation}
I\rightarrow E_{\theta^\star}(I)\rightarrow T_{\theta^\star}(I)
\rightarrow\mathrm{OrbitFormer}_{\theta^\star}(m=1)\rightarrow z.
\end{equation}
Evaluation performs cosine ranking only. We do not fit PCA, whitening, Procrustes transforms, clustering, target-domain normalization statistics, gallery calibration, test-time adaptation, or target-domain fine-tuning.

\section{Datasets and Evaluation Protocol}
\subsection{Per-dataset target-satellite-free development}
The late-arriving-gallery restriction is a \emph{task-specific} data-access rule. OCL may use grouped UAV training images from the benchmark being developed for, but it may not use that benchmark's satellite imagery for task-specific learning, validation, hyperparameter selection, or checkpoint selection. Accordingly:
\begin{itemize}[leftmargin=*]
    \item \textbf{University-1652} is the primary benchmark. Its source training split contains 701 locations, each with 54 observed UAV views. The fixed development split uses 631 scenes for optimization and 70 for source-only validation. Evaluation contains 37,855 drone queries and 951 satellite gallery images.
    \item \textbf{SUES-200} is evaluated separately at 150, 200, 250, and 300\,m. OCL is trained separately for each height using that height's UAV training split under the same target-satellite-free rule. Each height contains 120 source training scenes for geometry construction; all 120 complete orbit construction. The test protocol contains 4,000 UAV queries and 200 satellite gallery images per height.
    \item \textbf{IRVL328} uses infrared UAV source images and a visible satellite gallery. Its UAV training split contains 200 source scenes. SfM and 3DGS assets are available for all 200, while 187 pass the configured orbit/camera-matching criterion and are retained for OCL. The test set contains 7,040 infrared UAV queries and 328 visible satellite gallery images.
\end{itemize}
Source and test query scene IDs are disjoint in all three benchmark families. SUES-200 and IRVL328 results should therefore be read as \emph{target-satellite-free evaluation across datasets}, not as University-1652-to-target zero-shot transfer.

\subsection{Metrics}
All quantitative tables use the drone-to-satellite direction. We report R@1 and mean Average Precision (mAP) in the main paper; additional R@5 and R@10 are included below where available. Under these evaluation protocols every UAV query has one relevant satellite image, so mAP and MRR are numerically identical.

\subsection{Weather evaluation}
Following the WeatherPrompt evaluation style~\cite{wen2025WeatherPrompt}, University-1652 drone queries are evaluated under Fog, Rain, Snow, Fog+Rain, Fog+Snow, Rain+Snow, Dark, Over-exposure, and Wind, with the satellite gallery unchanged. The same weather query sets are used for all compared methods. OCL receives no weather-specific training.

The transformation implementation uses the following presets. ``Over-exposure'' corresponds to the internal \texttt{light} transform.
\begin{table}[h]
\centering
\scriptsize
\begin{tabularx}{\linewidth}{@{}lX@{}}
\toprule
Condition & Generation setting\\
\midrule
Fog & CloudLayer, intensity mean 225, coarse scale 2, alpha multiplier .9, sparsity .9, density .5\\
Rain & Five sequential rain layers; drop-size ranges from (.05,.10) to (.10,.20), speed (.04,.06)\\
Snow & Five sequential snow layers; flake-size ranges from (.5,.8) to (.6,.9), speed (.007,.03)\\
Dark & BlendAlpha(.5, Add(100), Multiply(.2)) followed by brightness multiplier .2 and additive shift $[-30,-15]$\\
Over-exposure & Brightness multiplier 1.6 and additive shift $[0,30]$\\
Fog+Rain & Fog followed by two rain layers\\
Fog+Snow & Fog followed by two snow layers\\
Rain+Snow & Interleaved snow/rain sequence using the same single-effect parameter ranges\\
Wind & Motion blur with kernel size 15\\
\bottomrule
\end{tabularx}
\caption{Weather-query generation settings. Fixed augmenter seeds are used in the evaluation implementation; all compared methods use the same resulting query sets.}
\label{tab:supp_weather_params}
\end{table}

\section{Baseline Provenance}
All tabulated baseline numbers are evaluated under the common drone-to-satellite protocol. Table~\ref{tab:supp_provenance} records the model source used in our evaluation. Most baselines use released checkpoints or pretrained models with our common evaluator; EM-CVGL is the exception and is locally retrained with the released method implementation. No baseline number in the main comparison is silently substituted by an OCL-tuned target-domain fit.

\begin{table}[h]
\centering
\small
\begin{tabularx}{\linewidth}{@{}l l X@{}}
\toprule
Method & Model source & Evaluation provenance\\
\midrule
Instance Loss & released checkpoint & local common-protocol evaluation\\
FSRA & released checkpoint & local common-protocol evaluation\\
Sample4Geo & released checkpoint & local common-protocol evaluation\\
Safe-Net & released checkpoint & local common-protocol evaluation\\
MFRGN & released checkpoint & local common-protocol evaluation\\
WeatherPrompt & released checkpoint & local common-protocol evaluation\\
SpikeViMFormer & released checkpoint & local common-protocol evaluation\\
EigenPlaces & released/pretrained checkpoint & local common-protocol evaluation, no target fitting\\
AnyLoc & released/pretrained checkpoint & local common-protocol evaluation, no target fitting\\
SALAD & released/pretrained checkpoint & local common-protocol evaluation, no target fitting\\
DINOv2 MeanPool & released pretrained backbone & local direct-transfer evaluation\\
DINOv3 MeanPool & released pretrained backbone & local direct-transfer evaluation\\
UCVGL & released checkpoint & local common-protocol evaluation\\
DMNIL & official U1652 checkpoint & local common-protocol evaluation\\
EM-CVGL & locally retrained released code & 200-epoch official-code training, then common-protocol evaluation\\
VFM-Loc & zero-shot pretrained DINOv3 & local evaluation; no PCA or Procrustes in the reported main-table variant\\
\bottomrule
\end{tabularx}
\caption{Baseline provenance. Checkpoint hashes and preprocessing records are retained with the evaluation outputs.}
\label{tab:supp_provenance}
\end{table}

\section{Complete Comparison Tables}
\subsection{University-1652}
\begin{table}[h]
\centering
\small
\begin{tabular}{@{}llcc@{}}
\toprule
Method & Pub. & R@1 & mAP\\
\midrule
\multicolumn{4}{@{}l}{\textit{Satellite-supervised}}\\
Instance Loss~\cite{zheng2020dual} & MM'20 & 56.89 & 66.41\\
FSRA~\cite{dai2021transformer} & TCSVT'21 & \textbf{98.65} & \textbf{99.29}\\
Sample4Geo~\cite{deuser2023sample4geo} & ICCV'23 & 92.11 & 94.57\\
Safe-Net~\cite{lin2024self} & TIP'24 & 80.42 & 85.89\\
MFRGN~\cite{wang2024mfrgn} & MM'24 & 93.55 & 95.44\\
WeatherPrompt~\cite{wen2025WeatherPrompt} & NeurIPS'25 & 82.40 & 87.27\\
SpikeViMFormer~\cite{chen2026efficient} & TCSVT'26 & 87.91 & 91.89\\
\midrule
\multicolumn{4}{@{}l}{\textit{Satellite-assisted weakly/self-supervised}}\\
EM-CVGL~\cite{li2024learning} & TGRS'24 & 70.22 & 79.42\\
UCVGL~\cite{li2024unleashing} & CVPR'24 & 36.59 & 44.89\\
DMNIL~\cite{chen2026without} & TNNLS'26 & \textbf{89.35} & \textbf{92.61}\\
\midrule
\multicolumn{4}{@{}l}{\textit{Gallery-blind target-satellite-free}}\\
EigenPlaces~\cite{berton2023eigenplaces} & ICCV'23 & 4.60 & 7.86\\
AnyLoc~\cite{keetha2023anyloc} & RA-L'23 & 17.08 & 25.49\\
SALAD~\cite{izquierdo2024optimal} & CVPR'24 & 8.43 & 14.34\\
\textbf{\ours (final main run)} & -- & \textbf{57.85} & \textbf{68.44}\\
\bottomrule
\end{tabular}
\caption{University-1652 drone-to-satellite retrieval (\%). Boldface is within each data-access category.}
\label{tab:supp_u1652}
\end{table}

\subsection{SUES-200 and IRVL328}
\begin{table*}[h]
\centering
\scriptsize
\setlength{\tabcolsep}{2.5pt}
\begin{tabular}{@{}llcccccccccc@{}}
\toprule
& & \multicolumn{8}{c}{SUES-200} & \multicolumn{2}{c}{IRVL328}\\
\cmidrule(lr){3-10}\cmidrule(lr){11-12}
Method & Pub. & \multicolumn{2}{c}{150\,m} & \multicolumn{2}{c}{200\,m} & \multicolumn{2}{c}{250\,m} & \multicolumn{2}{c}{300\,m} & \multicolumn{2}{c}{IR$\rightarrow$Vis.}\\
& & R@1&mAP&R@1&mAP&R@1&mAP&R@1&mAP&R@1&mAP\\
\midrule
\multicolumn{12}{@{}l}{\textit{Satellite-supervised}}\\
Instance Loss~\cite{zheng2020dual} & MM'20 & 23.77&38.12&30.55&43.30&31.95&44.32&33.90&45.26&4.66&10.66\\
FSRA~\cite{dai2021transformer} & TCSVT'21 & 51.73&64.80&61.58&73.33&68.00&78.30&72.72&81.61&11.63&21.38\\
Sample4Geo~\cite{deuser2023sample4geo} & ICCV'23 & 70.58&\textbf{79.12}&\textbf{81.53}&\textbf{87.71}&\textbf{86.58}&\textbf{91.25}&\textbf{90.12}&\textbf{93.60}&16.53&28.29\\
Safe-Net~\cite{lin2024self} & TIP'24 & 60.98&71.82&71.43&80.04&79.17&85.49&81.60&87.23&\textbf{23.95}&\textbf{34.78}\\
MFRGN~\cite{wang2024mfrgn} & MM'24 & 66.17&75.59&78.05&84.67&85.55&89.75&87.55&91.15&18.95&28.52\\
WeatherPrompt~\cite{wen2025WeatherPrompt} & NeurIPS'25 & 53.42&64.91&65.67&75.86&74.02&82.73&80.65&87.47&15.24&26.22\\
SpikeViMFormer~\cite{chen2026efficient} & TCSVT'26 & \textbf{70.60}&78.91&80.20&86.22&84.75&89.26&87.15&90.99&18.34&29.02\\
\midrule
\multicolumn{12}{@{}l}{\textit{Satellite-assisted weakly/self-supervised}}\\
EM-CVGL~\cite{li2024learning} & TGRS'24 & 61.50&72.46&72.80&81.87&81.67&88.12&86.78&91.44&10.01&21.49\\
UCVGL~\cite{li2024unleashing} & CVPR'24 & 21.10&31.38&30.03&40.56&35.05&45.50&38.75&49.41&9.15&15.82\\
DMNIL~\cite{chen2026without} & TNNLS'26 & \textbf{76.95}&\textbf{84.41}&\textbf{86.52}&\textbf{91.41}&\textbf{91.12}&\textbf{94.54}&\textbf{93.97}&\textbf{96.47}&\textbf{35.40}&\textbf{49.78}\\
\midrule
\multicolumn{12}{@{}l}{\textit{Gallery-blind target-satellite-free}}\\
EigenPlaces~\cite{berton2023eigenplaces} & ICCV'23 & 12.15&20.03&15.40&24.01&18.52&27.17&21.02&29.01&3.03&6.90\\
AnyLoc~\cite{keetha2023anyloc} & RA-L'23 & 34.83&48.09&40.98&53.46&46.75&58.64&48.63&60.77&5.33&13.20\\
SALAD~\cite{izquierdo2024optimal} & CVPR'24 & 15.57&24.12&21.62&30.30&25.55&34.93&30.50&40.13&3.86&9.09\\
DINOv2 MeanPool~\cite{oquab2024dinov} & TMLR'24 & \textbf{43.95}&\textbf{58.27}&49.63&63.72&59.60&71.02&65.92&75.96&7.84&17.82\\
DINOv3 MeanPool~\cite{simeoni2025dinov3} & arXiv'25 & 38.40&52.21&49.33&63.54&60.88&72.31&69.17&79.13&4.84&13.62\\
\textbf{\ours} & -- & 38.65&53.56&\textbf{53.62}&\textbf{66.17}&\textbf{62.68}&\textbf{73.98}&\textbf{69.33}&\textbf{79.47}&\textbf{11.79}&\textbf{26.42}\\
\bottomrule
\end{tabular}
\caption{Target-satellite-free evaluation across datasets (\%). OCL is trained separately from the corresponding UAV training split for each SUES altitude and for IRVL328; these rows are not University-1652-to-target zero-shot transfer.}
\label{tab:supp_multidataset}
\end{table*}

\subsection{Weather robustness}
\begin{table*}[h]
\centering
\scriptsize
\setlength{\tabcolsep}{2.3pt}
\begin{tabular}{@{}lccccccccc@{}}
\toprule
Method & Fog & Rain & Snow & F+R & F+S & R+S & Dark & Over-exp. & Wind\\
\midrule
\multicolumn{10}{@{}l}{\textit{Satellite-supervised}}\\
Instance Loss~\cite{zheng2020dual} & 0.29/2.03 & 1.13/3.34 & 1.68/4.13 & 0.05/0.74 & 0.18/0.95 & 1.00/2.82 & 0.01/0.62 & 4.95/10.20 & 13.84/22.43\\
FSRA~\cite{dai2021transformer} & 31.46/42.04 & 24.03/31.55 & 15.45/24.24 & 3.36/6.47 & 2.20/4.91 & 18.97/27.39 & 2.43/5.02 & 63.86/72.88 & \textbf{91.62/94.85}\\
Sample4Geo~\cite{deuser2023sample4geo} & 87.39/91.48 & 53.62/62.20 & 25.99/37.08 & 44.68/53.80 & 7.88/14.86 & 24.06/34.16 & 33.28/40.77 & 65.42/73.90 & 82.08/87.45\\
Safe-Net~\cite{lin2024self} & 35.76/46.15 & 23.12/30.92 & 13.92/21.05 & 10.58/16.83 & 2.86/6.01 & 15.67/22.90 & 13.11/19.60 & 43.33/53.40 & 63.32/71.65\\
MFRGN~\cite{wang2024mfrgn} & \textbf{88.17/91.79} & 50.39/58.73 & 35.11/45.16 & 45.72/54.59 & 19.85/27.72 & 35.09/44.41 & 23.80/30.03 & 70.88/78.03 & 86.05/90.26\\
WeatherPrompt~\cite{wen2025WeatherPrompt} & 80.75/86.05 & \textbf{79.90/85.40} & \textbf{76.45/82.72} & \textbf{78.59/84.38} & \textbf{72.42/79.47} & \textbf{77.71/83.69} & \textbf{68.69/76.03} & \textbf{73.58/80.51} & 75.78/82.22\\
SpikeViMFormer~\cite{chen2026efficient} & 69.27/77.53 & 37.57/46.82 & 13.70/22.45 & 23.14/31.59 & 3.42/6.72 & 16.87/24.98 & 21.50/28.69 & 65.46/74.51 & 60.34/69.50\\
\midrule
\multicolumn{10}{@{}l}{\textit{Satellite-assisted weakly/self-supervised}}\\
EM-CVGL~\cite{li2024learning} & 60.91/71.65 & 47.53/59.04 & \textbf{38.70/49.74} & 39.61/51.19 & \textbf{24.56/34.80} & \textbf{33.88/44.52} & \textbf{44.19/54.81} & 57.51/68.74 & \textbf{58.49/69.39}\\
UCVGL~\cite{li2024unleashing} & 32.67/40.70 & 17.17/23.23 & 4.88/9.07 & 15.67/21.25 & 1.74/4.68 & 8.42/13.16 & 20.82/27.24 & 29.62/37.60 & 13.85/20.04\\
DMNIL~\cite{chen2026without} & \textbf{85.90/90.12} & \textbf{57.77/65.78} & 21.87/30.38 & \textbf{55.25/63.25} & 8.65/13.95 & 31.74/40.87 & 42.01/49.29 & \textbf{76.56/82.79} & 52.88/61.60\\
\midrule
\multicolumn{10}{@{}l}{\textit{Gallery-blind target-satellite-free}}\\
EigenPlaces~\cite{berton2023eigenplaces} & 3.85/6.91 & 1.77/3.97 & 2.74/5.22 & 2.11/4.57 & 2.37/4.71 & 1.85/4.12 & 3.12/5.81 & 3.33/6.24 & 0.71/2.00\\
AnyLoc~\cite{keetha2023anyloc} & 28.98/39.93 & 16.89/26.47 & 20.08/29.76 & 15.10/24.41 & 13.99/22.47 & 14.29/22.99 & 16.38/24.14 & 24.34/34.63 & 22.44/32.66\\
SALAD~\cite{izquierdo2024optimal} & 8.60/14.76 & 4.85/9.34 & 6.07/11.33 & 5.16/9.91 & 5.42/10.22 & 4.48/8.91 & 5.19/9.94 & 5.98/11.19 & 6.05/11.30\\
DINOv2 MeanPool~\cite{oquab2024dinov} & 26.29/37.59 & 23.34/33.35 & 18.18/27.20 & 20.41/30.03 & \textbf{15.31/23.88} & 17.93/26.50 & 20.93/30.94 & 23.22/33.90 & 27.97/38.69\\
DINOv3 MeanPool~\cite{simeoni2025dinov3} & 28.59/40.86 & 19.38/29.12 & 18.15/27.58 & 16.43/25.71 & 13.08/21.64 & 14.54/23.20 & 19.40/29.04 & 25.51/36.66 & 26.91/38.42\\
\textbf{\ours} & \textbf{44.32/55.91} & \textbf{29.21/39.71} & \textbf{22.27/32.29} & \textbf{23.02/33.02} & 13.41/21.46 & \textbf{20.20/29.59} & \textbf{26.92/36.40} & \textbf{38.90/50.01} & \textbf{43.87/55.12}\\
\bottomrule
\end{tabular}
\caption{Weather robustness on University-1652, R@1/mAP (\%). The same weather query sets are used for all methods. Bold values indicate the best within each satellite-access category.}
\label{tab:supp_weather}
\end{table*}

\section{Ablation Details}
\subsection{Variant definitions}
\begin{table}[h]
\centering
\scriptsize
\begin{tabularx}{\linewidth}{@{}c l c c c X c@{}}
\toprule
ID & Configuration & Distill & Student & Teacher obs. & Teacher target & Validation queries\\
\midrule
M1 & Mean observed-only prototype & Yes & observed only & 54 observed & mean of singleton descriptors & observed only\\
M2 & + augmentation control & Yes & observed only & 54 observed + 16 fixed aug. & mean of singleton descriptors & observed only\\
M3 & + canonical completion & Yes & observed only & 54 observed + 16 3DGS & mean of singleton descriptors & observed only\\
M4 & Real-prototype NCE & No & observed only & 54 observed & mean of singleton descriptors & observed only\\
M5 & Joint real-orbit teacher & Yes & observed only & 54 observed & learned joint aggregation & observed + closure\\
M6 & + one rendered proxy & Yes & mixed & 54 observed + 1 render & learned joint aggregation & observed + closure\\
M7 & + 16-view completion & Yes & mixed & 54 observed + 16 renders & learned joint aggregation & observed + closure\\
M8 & Mean completed-orbit prototype & Yes & mixed & 54 observed + 16 renders & mean of singleton descriptors & observed + closure\\
\bottomrule
\end{tabularx}
\caption{Detailed M1--M8 configurations. ``Mixed'' denotes the 0.7 observed-only / 0.3 observed-plus-rendered pool policy.}
\label{tab:supp_variants}
\end{table}

M1--M3 were independently retrained as controlled teacher-context comparisons. M2 and M3 are observation-count matched: each adds 16 teacher observations, while the student sees only the original observed images and validation queries remain observed-only. M7 and M8 have matching saved training settings except for teacher-target aggregation and therefore form the cleanest teacher-aggregation comparison.

M1 and M4 provide supporting evidence about cosine prototype distillation. Their saved data split, initialization checkpoint, seed, training length, batch settings, learning rates, observation set, sampling schedule, teacher aggregation, EMA, memory settings, validation protocol, augmentation policy, and NCE settings match; the effective configured difference is the addition of $\mathcal L_{\mathrm{distill}}$ in M1. However, the two historical runs have different recorded training-code hashes and the historical source snapshots are unavailable for a source-level diff. We therefore treat M1-versus-M4 as \emph{configuration-matched supporting evidence}, not as a guaranteed strict one-variable experiment.

\subsection{Full fixed-seed retrieval metrics}
\begin{table}[h]
\centering
\small
\begin{tabular}{@{}lccccc@{}}
\toprule
Variant & R@1 & R@5 & R@10 & mAP & MRR\\
\midrule
M1 & 50.93 & 75.84 & 83.46 & 62.29 & 62.29\\
M2 & 51.49 & 76.11 & 83.56 & 62.76 & 62.76\\
M3 & 53.05 & 77.50 & 84.90 & 64.27 & 64.27\\
M4 & 49.71 & 74.65 & 82.09 & 61.11 & 61.11\\
M5 & 51.01 & 75.71 & 83.50 & 62.33 & 62.33\\
M6 & 53.81 & 78.30 & 85.52 & 64.97 & 64.97\\
M7 & 56.02 & 79.59 & 86.44 & 66.80 & 66.80\\
M8 & 57.84 & 81.00 & 87.43 & 68.44 & 68.44\\
\midrule
Final main U1652 run & 57.85 & 80.98 & 87.42 & 68.44 & 68.44\\
\bottomrule
\end{tabular}
\caption{University-1652 fixed-seed retrieval metrics (\%). M1--M8 are the ablation runs; the final main result is a distinct source-selected run.}
\label{tab:supp_full_ablation_metrics}
\end{table}

\subsection{Augmentation control}
M2 adds 16 fixed conventional augmentation observations per scene. Sixteen distinct source images are selected without replacement. Each augmented observation uses a deterministic per-scene seed; crop area is sampled from $[0.65,1)$ while preserving aspect ratio, crop location is uniformly sampled, the crop is resized with bicubic interpolation, horizontal flip is applied with probability 0.5, and brightness/contrast/saturation factors are independently sampled from $[0.8,1.2)$. These 16 augmented images are separate teacher-memory entries. They are fixed across epochs and are never sampled by the M2 student. M3 replaces exactly these 16 teacher entries by 16 canonical 3DGS renders, leaving the student and source-validation query protocol unchanged.

\subsection{Partial-orbit robustness}
The final OCL encoder supports variable-cardinality inputs. We construct a fixed nested manifest with $m\in\{1,2,4,8,16\}$ UAV observations per test location. Each location contributes one place-balanced multi-view query descriptor; the same subset is nested as $m$ increases.
\begin{table}[h]
\centering
\small
\begin{tabular}{@{}lccccc@{}}
\toprule
Views $m$ & 1 & 2 & 4 & 8 & 16\\
\midrule
R@1 (\%) & 58.06 & 63.05 & 65.05 & 65.62 & 67.19\\
mAP (\%) & 68.06 & 72.73 & 74.87 & 75.30 & 76.76\\
\bottomrule
\end{tabular}
\caption{Place-balanced retrieval as more UAV query observations become available. This diagnostic uses one fixed nested subset per location; its $m=1$ protocol therefore differs from the standard 37,855-query image-level benchmark.}
\label{tab:supp_partial_orbit}
\end{table}
Performance rises monotonically with additional observations, with the largest gain from one to two views. The single-image case remains the deployment default.

\section{Geometry Diagnostics}
\label{sec:geometry_diag}
\subsection{Source-side completion across datasets}
The geometry audit in Table~\ref{tab:supp_geometry_success} covers the current source assets used to construct OCL training orbits. ``Attempted'' counts source-training scenes, not test/query scenes. Test query scene IDs have zero overlap with the source reconstruction IDs.
\begin{table}[h]
\centering
\small
\begin{tabular}{@{}lrrrrr@{}}
\toprule
Dataset & Attempted & SfM & 3DGS & 16-view closure & Closure rate\\
\midrule
University-1652 & 701 & 701 & 701 & 701 & 100\%\\
SUES-150 & 120 & 120 & 120 & 120 & 100\%\\
SUES-200 & 120 & 120 & 120 & 120 & 100\%\\
SUES-250 & 120 & 120 & 120 & 120 & 100\%\\
SUES-300 & 120 & 120 & 120 & 120 & 100\%\\
IRVL328 & 200 & 200 & 200 & 187 & 93.5\%\\
\bottomrule
\end{tabular}
\caption{Current source-side reconstruction and orbit-completion assets. All IRVL328 scenes possess SfM and 3DGS assets; its 13 rejected scenes fail the subsequent orbit/camera-matching threshold rather than SfM or 3DGS itself.}
\label{tab:supp_geometry_success}
\end{table}

For IRVL328, 183 of the retained scenes match all 55 source cameras and four retained scenes match 51--54 of 55 cameras. The 13 rejected scene IDs are
\begin{quote}\small
0003, 0007, 0026, 0051, 0070, 0082, 0086, 0100, 0108, 0130, 0151, 0180, 0198.
\end{quote}
Their source-image-to-reconstructed-camera match ratios fall below the configured 0.9 threshold. Such scenes are dropped; the training code does not use a real-only fallback for failed closure scenes.

\subsection{University-1652 held-out rendering benchmark}
A separate 30-scene benchmark measures rendering fidelity and preprocessing cost. Scenes are sampled across reconstruction-complexity tertiles defined by historical COLMAP sparse-point count, while deliberately retaining difficult low-registration cases; the final distribution is 12/9/9 across the three tertiles. For every selected scene, eight views are held out from Gaussian fitting. These held-out images remain available to the preceding all-view COLMAP reconstruction and sparse initialization, so this is a held-out \emph{appearance/rendering} diagnostic given recovered source geometry, not a fully held-out SfM experiment.

\begin{table}[h]
\centering
\small
\begin{tabular}{@{}lc@{}}
\toprule
Diagnostic & Value\\
\midrule
Held-out PSNR & 26.39 dB\\
Held-out SSIM & 0.8669\\
Held-out LPIPS & 0.0830\\
\midrule
COLMAP, mean & 41.78 s / scene\\
Undistortion, mean & 0.25 s / scene\\
3DGS optimization, mean & 95.53 s / scene\\
Canonical rendering, mean & 4.01 s / scene\\
Primary pipeline, mean & 141.56 s / scene\\
Estimated 701-scene total & 27.56 h\\
\bottomrule
\end{tabular}
\caption{University-1652 30-scene reconstruction benchmark. The 701-scene total is a linear extrapolation of the sampled stage means, not a direct end-to-end timing of all 701 scenes.}
\label{tab:supp_geometry_quality}
\end{table}

The estimated total is $141.559\times701/3600\approx27.56$ hours. This cost is paid once while constructing source training assets and is separate from representation-learning runtime. Individual 3DGS views may still contain holes, floating geometry, incorrect occlusion, or view-dependent texture artifacts; OCL uses them only as training-time context.

\section{Efficiency Evidence}
The final per-dataset OCL models contain 123.46M trainable parameters during representation learning and produce 768-D descriptors. Table~\ref{tab:supp_runtime} reports wall-clock runtimes recorded by each training run on a single NVIDIA GeForce RTX 4090 D. The recorded timer includes model/data preparation performed within the run, memory initialization and refresh, prototype-bank construction, representation optimization, source validation, and checkpoint/log writing. It excludes the separate SfM/3DGS asset-construction stage reported above.

\begin{table}[h]
\centering
\small
\begin{tabular}{@{}lc@{}}
\toprule
Run & Recorded wall-clock runtime\\
\midrule
University-1652 final main run & 3 h 48 min\\
SUES-150 & 2 h 49 min\\
SUES-200 & 2 h 50 min\\
SUES-250 & 1 h 30 min\\
SUES-300 & 1 h 08 min\\
IRVL328 & 2 h 08 min\\
University-1652 development M8 run & 1 h 22 min\\
\bottomrule
\end{tabular}
\caption{Recorded representation-learning runtimes (20 epochs, one RTX 4090 D). The final University-1652 run and the development M8 run have different run scopes and should not be directly interpreted as identical timing benchmarks.}
\label{tab:supp_runtime}
\end{table}

For the final University-1652 run, the recorded PyTorch peak allocated GPU memory is approximately 3.21 GB. We do not report per-image inference latency because a controlled timing benchmark with documented warmup, synchronization, precision, and repetition settings was not retained with the evaluation artifacts. At deployment, inference nevertheless uses only the selected student image encoder; none of the source geometry pipeline, token memory, or prototype bank remains in the retrieval path.

\section{Qualitative Retrieval}
The qualitative cases are regenerated from the final University-1652 main checkpoint and its stored per-query descriptors rather than from an older ablation model. We use deterministic selection rules: the correct query with the largest hard margin, a correct query near the median margin, the error with the most negative hard margin, and the first erroneous query in query order. Table~\ref{tab:supp_qual_cases} records the selected cases.
\begin{table}[h]
\centering
\small
\begin{tabular}{@{}llll@{}}
\toprule
Case & Query identity & Correct rank & Hard margin\\
\midrule
success\_01 & place 0284 & 1 & +0.396\\
success\_02 & place 0725 & 1 & +0.087\\
failure\_01 & place 0395 & 606 & -0.720\\
failure\_02 & place 0000, image-02 & 3 & -0.009\\
\bottomrule
\end{tabular}
\caption{Selection provenance for the canonical qualitative examples.}
\label{tab:supp_qual_cases}
\end{table}

\begin{figure*}[h]
\centering
\setlength{\tabcolsep}{2pt}
\begin{tabular}{c c c c c c}
& Query & Ground truth & Top-1 & Top-2 & Top-3\\
\texttt{success\_01} & \caseimg{success_01/query.png} & \caseimg{success_01/gt.png} & \caseimg{success_01/topk_01.png} & \caseimg{success_01/topk_02.png} & \caseimg{success_01/topk_03.png}\\
\texttt{success\_02} & \caseimg{success_02/query.png} & \caseimg{success_02/gt.png} & \caseimg{success_02/topk_01.png} & \caseimg{success_02/topk_02.png} & \caseimg{success_02/topk_03.png}\\
\texttt{failure\_01} & \caseimg{failure_01/query.png} & \caseimg{failure_01/gt.png} & \caseimg{failure_01/topk_01.png} & \caseimg{failure_01/topk_02.png} & \caseimg{failure_01/topk_03.png}\\
\texttt{failure\_02} & \caseimg{failure_02/query.png} & \caseimg{failure_02/gt.png} & \caseimg{failure_02/topk_01.png} & \caseimg{failure_02/topk_02.png} & \caseimg{failure_02/topk_03.png}\\
\end{tabular}
\caption{Representative successes and failures from the final University-1652 OCL checkpoint. Rows are selected by deterministic retrieval-margin rules rather than hand-picked visual preference.}
\label{fig:supp_qualitative}
\end{figure*}

\section{Limitations and Broader Impact}
\paragraph{Source geometry.}
OCL assumes grouped multi-view source observations with enough overlap and appearance consistency for SfM and orbit construction. Weak texture, dynamics, severe illumination changes, missing camera registrations, or cross-modal appearance can reduce the quality or availability of the privileged geometry. IRVL328 illustrates this limitation: all 200 source scenes possess SfM/3DGS assets, but only 187 satisfy the configured orbit camera-match threshold. Failed scenes are dropped rather than repaired at representation-training time.

\paragraph{Finite completion.}
The 16-view closure is an engineered finite context set, not a complete model of all possible viewpoints. 3DGS renderings can contain geometric holes, floating structure, incorrect occlusions, or appearance artifacts, and full-to-partial distillation cannot recover scene evidence that is physically absent from a single query image.

\paragraph{Scope of ``target-satellite-free.''}
The claim concerns \emph{task-specific} access to the target satellite domain. OCL initializes from a generic DINOv3 backbone, whose large-scale pretraining should not be interpreted as guaranteed absence of aerial imagery. The experimental restriction is that no target benchmark satellite image participates in OCL parameter learning, source validation, hyperparameter selection, checkpoint selection, or target-domain statistical fitting.

\paragraph{Dual use.}
Visual geo-localization can support navigation, robotics, emergency response, and mapping, but can also contribute to surveillance and sensitive location tracking. Deployments should consider authorization boundaries, access control, data governance, and the handling of UAV observations and geotagged satellite galleries.

\section{Reproducibility Summary}
The main experimental claims can be traced to fixed source splits, stored source-selected checkpoints, per-query retrieval outputs, saved weather-query evaluations, source geometry manifests, and baseline protocol records. The ablation section is intentionally described as fixed-seed evidence rather than a multi-seed uncertainty study. The final deployed model is the selected student; EMA teacher state, source geometry, rendered views, token memory, and prototype banks are training-only artifacts.

{\small
\bibliographystyle{ieeenat_fullname}
\bibliography{references}
}